\documentclass{article}

\usepackage{microtype}
\usepackage{graphicx}
\usepackage{subcaption}
\usepackage{booktabs} 

\usepackage{hyperref}

\usepackage[accepted]{icml2026}

\usepackage{amsmath}
\usepackage{amssymb}
\usepackage{mathtools}
\usepackage{amsthm}
\usepackage{multirow}
\usepackage{makecell}

\usepackage[capitalize,noabbrev]{cleveref}

\theoremstyle{plain}

\theoremstyle{definition}

\theoremstyle{remark}

\usepackage[textsize=tiny]{todonotes}

\icmltitlerunning{Graph is a Natural Regularization: Revisiting Vector Quantization for Graph Representation Learning}

\begin{document}

\twocolumn[
  \icmltitle{Graph is a Natural Regularization: Revisiting Vector Quantization for Graph Representation Learning}



  \icmlsetsymbol{equal}{*}

  \begin{icmlauthorlist}
    \icmlauthor{Zian Zhai}{yyy}
    \icmlauthor{Fan Li}{yyy}
    \icmlauthor{Xingyu Tan}{yyy}
    \icmlauthor{Xiaoyang Wang}{yyy}
    \icmlauthor{Wenjie Zhang}{yyy}
  \end{icmlauthorlist}

  \icmlaffiliation{yyy}{School of Computer Science and Engineering, University of New South Wales, Sydney, Australia}

  \icmlcorrespondingauthor{Fan Li}{fan.li8@unsw.edu.au}

  \icmlkeywords{Machine Learning, ICML}

  \vskip 0.3in
]
\newcommand{\myparagraph}[1]{\vspace{0.2mm} \noindent \textbf{#1}.}


\printAffiliationsAndNotice{}  

\begin{abstract}
Vector Quantization (VQ) has recently emerged as a promising approach for learning compressed and discrete representations for graph-structured data.
However, a fundamental challenge, i.e., codebook collapse, remains underexplored in the graph domain, significantly limiting the expressiveness and generalization of graph tokens.
In this paper, we present an empirical study and observe that codebook collapse consistently occurs when training VQ jointly with Graph Neural Networks under graph reconstruction tasks, even with mitigation strategies proposed in vision or language domains. 
Moreover, we provide a diagnosis of collapse from data and optimization perspectives, showing that collapse is associated with graph data properties such as feature redundancy and connectivity density, and is further reinforced by the training dynamics of deterministic hard assignment.
To address these issues, we propose RGVQ, a novel framework that integrates graph topology and feature similarity as explicit regularization signals to enhance codebook utilization and promote token diversity. 
RGVQ introduces soft assignments via Gumbel-Softmax reparameterization, ensuring that all codewords receive gradient updates. 
In addition, RGVQ incorporates a structure-aware contrastive regularization to penalize assigning the same token to dissimilar node pairs.
Extensive experiments demonstrate that RGVQ substantially improves codebook utilization and consistently boosts the performance of state-of-the-art graph VQ backbones across multiple downstream tasks, enabling more expressive and transferable graph token representations.
\end{abstract}

\section{Introduction}

In recent years, a discretization-based tokenization method, known as Vector Quantization (VQ), has attracted significant research attention for its effectiveness in generative modeling~\cite{van2017vqvae, caron2018eccv}.
VQ quantizes continuous latent representations into discrete clusters referred to as ``codewords" in a learnable codebook~\cite{zhang2023VQimage}.
These codewords are then trained to reconstruct the original data samples. 
By discretizing latent space, VQ provides an effective prior for learning disentangled features, and has achieved remarkable success in generative tasks, including image synthesis~\cite{ramesh2021zero, chang2023muse,li2024autoregressive}, speech generation~\cite{dhariwal2020jukebox,zhang2024codebook}, and language models~\cite{liu2025llmvq,van2024gptvq}.

Motivated by these successes, recent efforts have begun to explore the extension of VQ to graphs for scalable and versatile graph tokenization.
First, discretizing graphs into VQ tokens enables compact graph compression, substantially reducing the memory and computation overhead during inference~\cite{yang2023vqgraph, luo2024nodeid}.
Second, VQ provides a natural mechanism for abstracting structural patterns into a reusable token vocabulary, analogous to the language tokens used in Large Language Models (LLMs), and offers a promising pathway toward Graph Foundation Models (GFMs)~\cite{wang2024gft}.
Third, VQ allows graphs to be serialized into token sequences, enabling sequence-based modeling with standard Transformer architectures that are widely adopted in NLP and vision, and eliminating the need for handcrafted inductive biases that are typically required in Graph Transformers~\cite{wang2025learning}.

Similar to VQ models in the vision and language domains, which are typically trained to reconstruct input samples jointly with encoders~\cite{navaneet2024compgs,deng2024autoregressive}, Graph VQ is likewise trained jointly trained with Graph Neural Networks (GNNs) under graph reconstruction objectives, including both node feature and edge reconstruction.
Nevertheless, through our empirical study, we observe that \textbf{codebook collapse} consistently occurs, even when applying mitigation strategies that are commonly used in other domains.
This refers to the phenomenon where most inputs are mapped to only a few codewords, leaving the majority underutilized~\cite{zhu2024softassign, zhang2023VQimage}.
As a result, only a limited number of tokens can be utilized during inference, leading to overly coarse representations and degradation in task performance. 
However, prior work focuses only on the performance of downstream tasks, without addressing this critical problem in the first place~\cite{wang2024gft,zeng2025hgvq,yang2023vqgraph}.
This consistent underutilization naturally raises a central question: \textit{What makes Graph VQ more prone to collapse?}

To answer this question, we diagnose the collapse from both \textit{data} and \textit{optimization} perspectives.
From the data perspective, we find that the severity of collapse is associated with typical graph properties such as feature redundancy and connectivity density. In particular, graphs with higher feature redundancy and denser connectivity are linked to more severe collapse, suggesting that intrinsic properties of graph data can exacerbate the issue.
From the optimization perspective, we analyze the training dynamics of deterministic VQ and show that hard assignment induces a self-reinforcing feedback loop:
frequently selected codewords receive more updates and become increasingly dominant, while rarely selected ones remain inactive, which limits utilization exploration and ultimately drives the system toward collapse.

Based on these insights, we propose \textbf{Regularized Graph Vector Quantization (RGVQ)}, a novel framework that integrates graph topology and feature similarity as explicit regularization signals to enhance codebook utilization. 
First, to break the self-reinforcing loops, RGVQ adopts the Gumbel-Softmax reparameterization to relax hard assignments into differentiable probability distributions, enabling the gradients to flow not only to the most likely codewords but also to the less probable candidates. 
Second, RGVQ leverages graph topology and feature similarity to regularize token assignment distributions, explicitly penalizing overly concentrated token utilization induced by graph redundancy. 
This regularization encourages nodes with similar features and local structures to share token distributions, while discouraging similar assignment among unrelated nodes.

Our contributions can be summarized as follows.
\begin{itemize}
    \item To the best of our knowledge, we provide a systematic empirical study of codebook collapse on graphs, benchmarking mitigation strategies from other domains and revealing collapse as a fundamental bottleneck in discrete graph token learning.
    \item We provide a collapse diagnosis from data and optimization perspectives, and identify that collapse is associated with graph redundancy and hard-assignment in deterministic VQs.
    \item We propose RGVQ to address graph data redundancy by structure-aware regularization and disrupts the self-reinforcing dynamics via stochastic quantization.
    \item We perform comprehensive experiments on state-of-the-art (SOTA) Graph VQ backbones, demonstrating that our proposed method improves codebook utilization and downstream performance, and serves as a flexible plug-in for learning graph tokens.
\end{itemize}

\section{Related Work}
\myparagraph{Vector Quantization} 
Vector Quantization (VQ) maps continuous inputs to discrete tokens in a codebook and has been widely used in image, video, and audio generation~\cite{chung2020vector, fifty2024restructuring, tang2022avqvc}.
This success has motivated efforts to extend VQ to graph data.
For example, VQ-GNN~\cite{ding2021vqgnn} and VQGraph~\cite{yang2023vqgraph} apply VQ for embedding compression, but their fully supervised training deviates from the original unsupervised training scheme of VQ~\cite{chen2021incremental,yu2021ortho}.
More recently, GFT pretrains VQ by reconstructing graph features to utilize the learned codebook as transferable vocabulary across tasks and domains~\cite{wang2024gft}. 
GQT employs residual VQ to tokenize graphs for vanilla transformers, alleviating manual architectural bias in graph transformers~\cite{wang2025learning}.
While both methods demonstrate promising applications of Graph VQ, they overlook the issue of codebook collapse, which undermines the generalization of learned tokens.
HQA-GAE introduces a hierarchical VQ, improving performance on graph tasks~\cite{zeng2025hgvq}.
Nevertheless, it does not effectively resolve the non-differentiability of VQ and lacks a formal analysis of codebook collapse.

\myparagraph{Collapse Mitigation} 
One of the most fundamental limitations of VQ is codebook collapse, wherein only a small fraction of codewords are used~\cite{zhang2024codebook, lu2023hierarchical}.
Various mitigation strategies have been explored.
Exponential Moving Average (EMA) is proposed to stabilize codebook updates~\cite{polyak1992acceleration, wu2019vector}.
Pretraining the encoder~\cite{zhao2024pretrainVQ} is proposed to mitigate embedding drift during training VQ.
In addition, codebook reset~\cite{zeghidour2021reset, williams2020hierarchical} periodically reinitializes inactive codewords with encoder embeddings.
Affine parameters~\cite{huh2023straightening, zhang1997improvement} introduce a learnable transformation to align encoder outputs with the codebook space.
Recently, SimVQ~\cite{zhu2024softassign} reparameterizes the code vectors through a linear transformation layer based on a learnable latent basis.
Although these mitigation strategies have been evaluated in image and speech domains, their performance on graph data remains underexplored.

\section{Preliminary}
\myparagraph{Graph Neural Network}
Graph Neural Networks (GNNs) learn the node representations by recursively aggregating features from neighbors, also known as message-passing~\cite{zhai2025sgpt, tan2026memotime, tan2026privgemo, li2025dhg}. Formally, the representation of node $v$ at the $l$-th layer is:
\begin{equation}
    \mathbf{h}^{(l)}_v = \text{AGG}(\{\mathbf{h}_u^{(l-1)}, u \in \mathcal{N}(v) \cup v\}, \phi^{(l)}),
\end{equation}

where $\mathbf{h}^{(0)}_v = \mathbf{x}_v$ is the initial node feature, $\mathcal{N}(v)$ is the neighbor set of node $v$, and $\phi^{(l)}$ is the parameters of the $l$-th layer of the GNN. 
The aggregation function $\text{AGG}(\cdot)$ combines the embedding of node $v$ and its neighbors, which is typically implemented as sum, mean, or max pooling~\cite{xu2026c2tctrainingfreeframeworkefficient,li2026fairness,zheng2026rethinkinggatingmechanismsparse}. 

\myparagraph{Deterministic VQ}
VQ maps continuous vectors into a finite set of discrete embeddings in the codebook~\cite{van2017vqvae}. 
Given a codebook $\mathbf{C} = [\mathbf{e}_1,...,\mathbf{e}_K ] \in \mathbb{R}^{K \times d}$ with each discrete codeword $\mathbf{e}_i \in \mathbb{R}^d$, a continuous input $\mathbf{h}_i \in \mathbb{R}^d$ is quantized as $\mathbf{z}_i$ with the nearest codeword $\mathbf{e}_k$ by:
\begin{equation}
    k = \arg\min_{j} \|\mathbf{h}_i - \mathbf{e}_j\|_2^2 = \arg\min_{j} \|\mathbf{h}_i - \delta_j\mathbf{C}\|_2^2,
\end{equation}
where $\delta_j \in \{0, 1\}^{1\times K}$ is the one-hot indicator vector with only the $j$-th element being 1.
To enable gradient propagation through the non-differentiable vector $\delta_j$, the Straight-Through Estimator (STE) is applied~\cite{bengio2013estimating}. 
During the backward process, the gradient of the quantized embedding $\mathbf{z}_i = \delta_j\mathbf{C}$ is copied to $\mathbf{h}_i$, which is denoted as
\begin{equation}
    \mathbf{z}_i = \text{sg}[\delta_j\mathbf{C} - \mathbf{h}_i] + \mathbf{h}_i, \quad \Rightarrow \frac{\partial \mathbf{z}_i}{\partial \mathbf{h}_i} = 1,
\end{equation}
where $\text{sg}[\cdot]$ denotes the stop-gradient operator, ensuring the gradient for one-hot selection $\delta_j\mathbf{C}$ is discarded during the backward process.
Finally, the learning objective is to reconstruct the input samples, with a codebook loss that pulls the quantized representations $\mathbf{Z} = [\mathbf{z}_1, \mathbf{z}_2, \dots, \mathbf{z}_N]$ toward the encoder outputs $\mathbf{H}= [\mathbf{h}_1, \mathbf{h}_2, \dots, \mathbf{h}_N]$, and a commitment loss that pulls the encoder outputs toward the quantized representations:
\begin{equation}
    \mathcal{L}_{\text{VQ}} = 
    \underbrace{\mathcal{L}_{\text{recon}}}_{\text{reconstruction loss}} + 
    \underbrace{\|\text{sg}[\mathbf{H}] - \mathbf{Z}\|^2_2}_{\text{codebook loss}} + 
    \beta \underbrace{\|\mathbf{H} - \text{sg}[\mathbf{Z}]\|^2_2}_{\text{commitment loss}}.
\label{eq: vqloss}
\end{equation}
For Graph VQ, the reconstruction task typically involves reconstructing the graph properties, i.e., node features and links~\cite{wang2024gft, wang2025learning, yang2023vqgraph}:
\begin{equation}
{ \mathcal{L}_{\text{recon}} = 
\underbrace{\frac{1}{N} \left\| \mathbf{X} - \hat{\mathbf{X}}\right\|^2_2}_{\text{feature reconstruction}} + 
\underbrace{ \left\|\mathbf{A} - \hat{\mathbf{A}} \right\|_2^2}_{\text{link reconstruction}}},
\end{equation}
where $\mathbf{A}\in \mathbb{R}^{N\times N}$ denotes the adjacency matrix, $N$ is the total number of nodes, and $\mathbf{X}\in \mathbb{R}^{N\times D}$ is the node feature matrix. The reconstructed feature matrix $\hat{\mathbf{X}} = g_{\theta _{1}}(\mathbf{Z})$ and the reconstructed adjacency matrix $\hat{\mathbf{A}} = g_{\theta _{2}}(\mathbf{Z})$ are generated by the decoders for the feature reconstruction and link reconstruction tasks, respectively~\cite{wang2024gft, zeng2025hgvq}.

\myparagraph{Metric for Codebook Utilization}
The extent of codebook collapse is measured by the codebook perplexity~\cite{takida2022sqvae, yan2024gaussian, zheng2023online}, which is defined as: 
\begin{equation}
P = \exp\left( - \sum_{k=1}^{K} p_k \log p_k \right),
\end{equation}
where $p_k$ denotes the probability of selecting the $k$-th codeword. 
A low perplexity indicates that only a few codewords dominate the assignments, reflecting a high degree of collapse. 
In contrast, a high perplexity suggests better utilization of the codebook capacity.
\begin{figure*}[t]
  \centering
\includegraphics[width=0.95\linewidth]{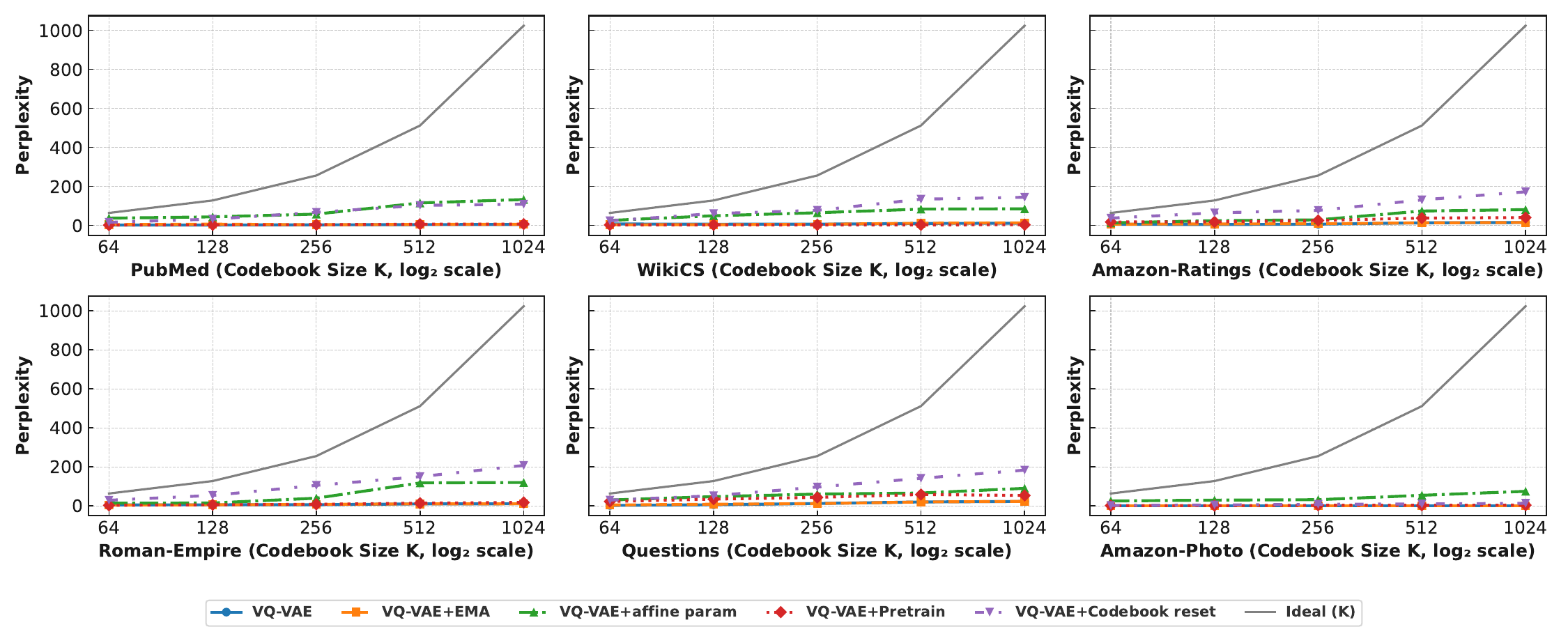}
\caption{Codebook perplexites on graph datasets. The black lines indicate the optimal perplexities, i.e., codebook size K.}
  \label{fig:study}
\end{figure*}

\section{Motivation}
Recent studies demonstrate the potential of Graph VQ; however, most existing
methods primarily focus on downstream performance, leaving codebook utilization
largely underexplored.
This limitation becomes a bottleneck that hinders the scalability of graph tokens
with different codebook size, thereby preventing flexible choices between
expressive and compressed tokenization.
Moreover, overly concentrated token assignments also impair downstream
performance, as shown in Section~\ref{Exp:abl}.
In this section, we conduct an empirical study of the codebook utilization and observe that codebook collapse occurs consistently during training, even when applying SOTA mitigation strategies from the language and vision domains. 
Furthermore, We diagnose this phenomenon from both data and optimization perspectives.
From a data perspective, we find that the severity of collapse correlates with the properties of graph data: higher feature redundancy and denser local connectivity are associated with lower codebook utilization.
From an optimization perspective, we show that deterministic hard assignment induces a self-reinforcing feedback loop, under which rarely selected codewords receive little update and become difficult to reactivate once they fall behind, thereby limiting effective exploration of the codebook.
These insights motivate the development of our method.

\subsection{Empirical Study}
We begin by investigating the codebook perplexity of Graph VQ on different graph datasets. 
Following the settings of prior work~\cite{ding2021vqgnn, wang2024gft, yang2023vqgraph}, we apply vanilla Graph VQ and its variants augmented with SOTA collapse mitigation methods, including EMA, codebook reset, pretrained encoder, and affine parameters. 
By default, orthogonal normalization~\cite{yu2021ortho} and cosine similarities~\cite{wang2024gft} 
are incorporated in all variants.
The implementation details and additional datasets can be found in Appendix~\ref{appendix:experiments} and~\ref{appendix:additional}, respectively.
From Figure \ref{fig:study}, we make the following observations:
\noindent
\textbf{(Ob. 1) }Codebook collapse is a systematic and severe issue in Graph VQ. 
Across all datasets, the perplexity of VQ remains far below the codebook capacity, and fails to grow proportionally with the increasing codebook size. 
\noindent
\textbf{(Ob. 2)} General mitigation strategies adopted from other domains only achieve marginal improvements and fail to fundamentally address codebook collapse in graphs. 
These findings reveal that codebook collapse is not merely incidental, but a systematic issue in Graph VQ. 

\subsection{Collapse Diagnosis}
\begin{figure}[h]
  \begin{subfigure}[t]{0.48\linewidth}
    \centering
    \includegraphics[width=\linewidth]{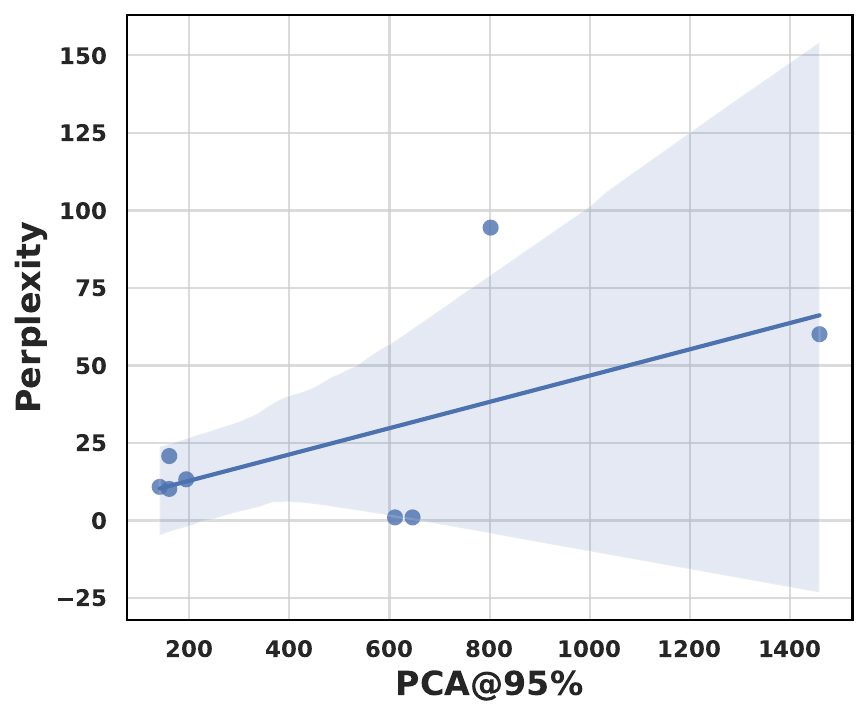}
    \caption{PCA@95\% vs. Perplexity}
  \end{subfigure}
  \begin{subfigure}[t]{0.48\linewidth}
    \centering
    \includegraphics[width=\linewidth]{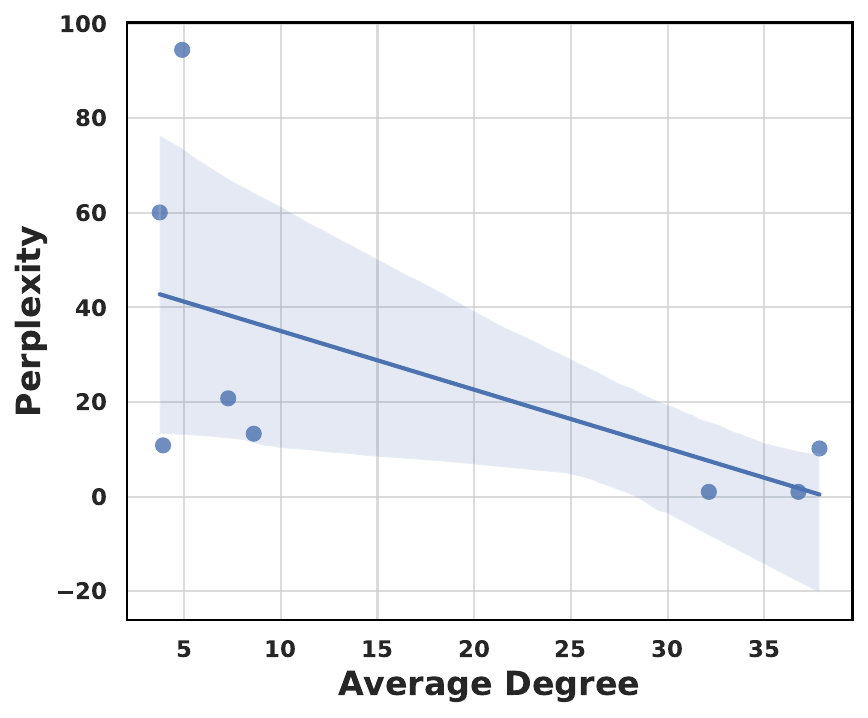}
    \caption{Avg. Degree vs. Perplexity}
  \end{subfigure}
  \caption{Correlation between graph properties and perplexity.}
  \label{fig:property_vs_perplexity1}
\end{figure}

\myparagraph{Graph Properties}
As codebook collapse consistently occurs in graphs, we hypothesize that the unique properties of graph data, i.e., inherent feature redundancy and non-i.i.d.\ nature, may contribute to this phenomenon.
To investigate this, we analyze two graph-level statistics that serve as proxies for these properties on investigated datasets. 
We consider:
(1) PCA@95\%, which quantifies feature redundancy by measuring the number of principal components needed to preserve 95\% of node feature variance~\cite{dong2022denoising, hou2023graphmae2}; and
(2) average node degree, which reflects local connectivity density. 
Higher degrees imply stronger dependencies between neighboring nodes, violating the i.i.d.\ assumption and serving as a simple proxy for the non-i.i.d.\ nature of graph data~\cite{yang2023vqgraph, wu2019net}.
From Figure~\ref{fig:property_vs_perplexity1}, we observe a positive correlation between PCA@95\% and codebook perplexity, and a negative correlation between average degree and perplexity. 
Lower PCA@95\% suggests higher feature redundancy, and higher average degree implies stronger local connectivity and non-i.i.d.\ characteristics, both associated with a greater tendency toward collapse.
Additionally, we provide dataset statistics (PCA@95, average degree) and measured codebook perplexity across 8 graph datasets in Appendix~\ref{appendix:additional}.

\myparagraph{Optimization Dynamics}
To complement the data-side diagnosis, we examine the optimization dynamics of deterministic VQ and show how hard assignment can create a positive feedback loop that drives codebook collapse.
In Graph VQ, the codebook $\mathbf{C}$ is updated only through the vocabulary loss~\cite{zhu2024softassign}, i.e., the second term in Equation~\ref{eq: vqloss}. 
The update is denoted as:
\begin{equation}
\mathbf{C}^{(t+1)} = \mathbf{C}^{(t)} 
- \eta\, \mathbb{E}_{\mathbf{h}_i} \left[ \delta_k^\top \delta_k\, \mathbf{C}^{(t)} \right]
+ \eta\, \mathbb{E}_{\mathbf{h}_i} \left[ \delta_k^\top\, \mathbf{h}_i \right],
\label{eq:codebook_update}
\end{equation}
where $\mathbf{h}_i$ is the embedding of node $v_i$, $\eta$ is the learning rate, and $\delta_k^\top \delta_k$ is the Kronecker delta matrix, defined as:
\begin{equation}
    (\delta_k^\top \delta_k)_{ij} = 
    \begin{cases}
1 & \text{if } i = j = k, \\
0 & \text{otherwise}.
\end{cases}
\end{equation}

\begin{figure*}[t]
  \centering
\includegraphics[width=0.7\linewidth]{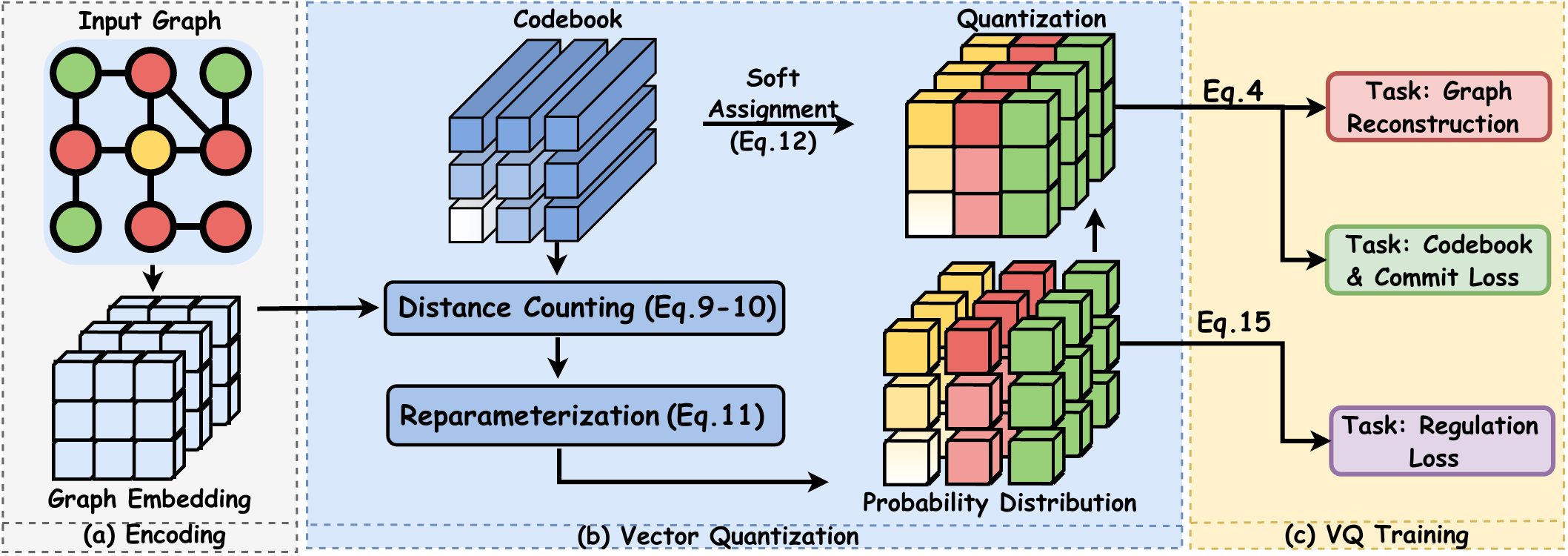}
\caption{Overall framework of RGVQ. Note: red nodes represent the positive set, while green nodes denote negative samples.}
  \label{fig:fram}
\end{figure*}
This condition indicates that if and only if when the expectation $\mathbb{E}_{\mathbf{h}_i}[\delta_k^\top \delta_k] = \mathbf{I},$ i.e., every token is selected with equal probability $\frac{1}{K}$, each codebook entry is updated during training.
In practice, according to the codebook loss term in Equation~\ref{eq: vqloss}, selected codewords are updated and pulled towards the distribution of the output of the GNN, i.e., $\mathbf{h}_i$.
On the other hand, the encoder outputs are simultaneously optimized towards the selected codewords via the commitment loss in Equation~\ref{eq: vqloss}.
This hard assignment and bidirectional attraction form a self-reinforcing ``cocoon effect,'' which not only locks the encoder into preferring codewords, but also suppresses any possibility of unused codewords, specifically $(\mathbf{I} - \mathbb{E}_{\mathbf{h}_i} [ \delta_k^\top \delta_k ]) \mathbf{C}$, being reactivated. 
\section{Methodology}
Our collapse diagnosis identifies two key contributors to codebook collapse: (i) graph data properties and (ii) the self-reinforcing training dynamics of deterministic VQ with hard assignment.
Accordingly, we propose RGVQ, a regularized Graph VQ framework consisting of two corresponding components, as illustrated in Figure~\ref{fig:fram}.
First, to break the optimization loop of deterministic VQ, RGVQ replaces hard assignments with differentiable assignment distributions among all codewords using Gumbel-Softmax reparameterization, enabling gradients to flow to all codewords proportionally to their assignment probabilities.
Second, to mitigate the effect of graph data redundancy, RGVQ leverages graph topology and feature similarity to regularize token assignment distributions, explicitly discouraging overly concentrated token usage from dissimilar nodes and promoting higher token diversity.
RGVQ can be added to the loss in Equation~\ref{eq: vqloss}.
Additionally, we provide the detailed sampling procedures and complexity in Appendix~\ref{appendix:complexity}.

\myparagraph{Gumbel-Softmax Reparameterization}
In deterministic VQ, the training dynamics of hard assignments prevent gradient backpropagation to unselected codewords, ultimately leaving them inactive and underutilized.
To address this issue, we adopt Gumbel-softmax reparameterization~\cite{roy2018soft1, sonderby2017soft2}, which replaces hard nearest-neighbor assignment with a differentiable soft selection.
Formally, given node embedding $\mathbf{h}_i$, we define its assignment logit $\boldsymbol{\pi}_i\in\mathbb{R}^K$ with each element computed by
\begin{equation}
\pi_{ik} = -\|\mathbf{h}_i - \mathbf{e}_k\|_2^2,\quad k=1,\ldots,K,
\label{eq:gumbel_logits}
\end{equation}
and the assignment probability vector $\mathbf{p}_i\in\mathbb{R}^K$ is denoted as:
\begin{equation}
\mathbf{p}_i = \text{Softmax}(\boldsymbol{\pi} _i).
\label{eq:gumble}
\end{equation}

Instead of using a non-differentiable $\text{argmax}$ over the distribution, we apply the Gumbel-Softmax trick to estimate a differentiable approximation of this hard assignment.
Specifically, the assignment distribution is perturbed with Gumbel noise and passed through a temperature-controlled softmax:
\begin{equation}
\tilde{\mathbf{p}}_i= \text{Softmax}_\tau \left( \log \mathbf{p}_i + \mathbf{g}_i \right),
\label{eq:gumbel-softmax}
\end{equation}
where $\mathbf{g}_i \in \mathbb{R}^K$ is a noise vector with i.i.d.\ entries $g_{ik}\sim\mathrm{Gumbel}(0,1)$, and $\tau$ is the temperature. 
Given this estimated distribution, the quantized embedding is computed as a weighted average over all codebook entries:
\begin{equation}
\tilde{\mathbf{z}}_i = \mathbf{C}^\top \tilde{\mathbf{p}}_i .
\end{equation}
Because $\tilde{\mathbf z}_i$ is a differentiable weighted sum of all codewords, gradients propagate to each $\mathbf e_k$ proportionally to $\tilde p_{ik}$, rather than only to the selected entry in deterministic VQ, thereby mitigating the aforementioned self-reinforcing loop.
During inference, the model reverts to deterministic hard assignment by selecting the codeword with the maximum logit $j = \arg\max_j \pi_{ij}$ and setting $\mathbf z_i = \mathbf e_{j}$.

\begin{algorithm*}[tb]
  \caption{Training procedure of RGVQ}
  \label{alg:rgvq}
  \begin{algorithmic}
    \STATE {\bfseries Input:} Encoder $f_\phi$, Decoder $g_\theta$, Codebook $\mathbf C=[\mathbf e_1, \dots, \mathbf e_K]$, Temperature $\tau$, commitment weight $\beta$.
    \STATE Initialize codebook $\mathbf{C}$ using K-means.
    \STATE Compute positive set $\mathcal{N}_P$ and negative set $\mathcal{N}_N$ for each node.
    \REPEAT
      \STATE Sample minibatch $\mathbf{x} \sim p_{\text{data}}$.
      \STATE Compute embeddings $\mathbf{h} \gets f_\phi(\mathbf{x})$.
      \STATE Compute logits $\pi_k \gets -\|\mathbf{h}-\mathbf{e}_k\|^2$ for $k=1,\ldots,K$.
      \STATE Assignment distribution $\mathbf{p} \gets \mathrm{Softmax}(\boldsymbol{\pi})$.
      \STATE Sample Gumbel noise $\mathbf{g}\in\mathbb{R}^K$ with i.i.d.\ $g_k\sim\mathrm{Gumbel}(0,1)$.
      \STATE Soft assignment $\tilde{\mathbf{p}} \gets \mathrm{Softmax}_\tau(\log \mathbf{p}+\mathbf{g})$.
      \STATE Soft quantization $\tilde{\mathbf{z}} \gets \mathbf{C}^\top \tilde{\mathbf{p}}$.
      \STATE Update parameters by minimizing
      \STATE \hspace{1em} $\mathcal{L} =
          \mathcal{L}_{\text{recon}}
          + \mathcal{L}_{\text{reg}}
          + \| \mathrm{sg}[\mathbf{H}] - \tilde{\mathbf{Z}} \|^2
          + \beta \| \mathbf{H} - \mathrm{sg}[\tilde{\mathbf{Z}}] \|^2.$
    \UNTIL{Converge}
  \end{algorithmic}
\end{algorithm*}
\myparagraph{Structure-Aware Regularization}
To mitigate feature redundancy and dense connectivity in graph data, we incorporate feature and structural similarities to regularize the token assignment distribution computed from Gumbel-Softmax Reparameterization, encouraging the model to avoid overuse of some specific codebook entries.
Our key insight is that collapse arises when nodes are spuriously mapped to the same tokens due to overly similar embeddings caused by dense connectivity or similar features. 
Therefore, we explicitly distinguish between similar and dissimilar node pairs based on both feature and local connectivity: similar nodes can exhibit more consistent assignment distributions, while dissimilar nodes should be discouraged from sharing similar distributions. 
Formally, given an anchor node $v$, we define:
\begin{itemize}
    \item \textbf{Positive set} $\mathcal{N}_P$: consists of $n$ sampled nodes that are either structurally or semantically similar to $v$. Specifically, $n$ positive nodes are sampled from the union of the following two candidate sets:
    (1) nodes directly connected to $v$; or 
    (2) the top $K$ feature-similar nodes to $v$. 
    Formally, the positive set is denoted as:
\begin{equation}
\begin{aligned}
    &\mathcal{N}_P =
    \left\{\, 
    u \;\middle|\; 
    (a_{uv}=1) \;\lor\;  \right.\\
    &
    \Big(
    u \in \operatorname{arg\,topk}_{u' \in \mathcal{V}}
    \text{sim}(\mathbf{x}_{u'}, \mathbf{x}_v)
    \Big)\Big\},
\end{aligned}
\end{equation}
    where $a_{uv} \in \mathbf{A}$ is the adjacency matrix, sim(·,·) is the similarity function, $\mathbf{x}_v$ is the feature of node $v$. We apply the cosine similarity as the similarity function.
    \item \textbf{Negative set} $\mathcal{N}_N$: 
consists of $n$ sampled nodes that are neither structurally connected nor semantically similar to $v$.
Formally, the negative set is defined as:
\begin{equation}
\begin{aligned}
    &\mathcal{N}_N =
    \left\{\, 
    u  \;\middle|\; 
    (a_{uv}=0) \;\land\;  \right.\\
    &
    \!\!\!\!\!\!\Big(
    u \notin \operatorname{arg\,topk}_{u' \in \mathcal{V}}
    \text{sim}(\mathbf{x}_{u'}, \mathbf{x}_v)
    \Big)
    \;\land\; (u \neq v)
    \,\Big\}.
\end{aligned}
\label{eq:negative}
\end{equation}
\end{itemize}
We encourage nodes in the positive set to have similar assignment distributions, while penalizing nodes in the negative set for having overlapping token distributions.
Formally, given two nodes $v_i$ and $v_j$ and their token assignment distributions $\tilde{\mathbf{p}}_i$ and $\tilde{\mathbf{p}}_j$, the distributions are regularized by an InfoNCE loss~\cite{you2021graph,wu2021rethinking}, which is defined as:
\begin{equation}
\mathcal{L}_i = -\log \frac{\sum_{j \in \mathcal{N}_P} \exp(\text{sim}(\tilde{\mathbf{p}}_i, \tilde{\mathbf{p}}_j))}{\sum_{j \in \{\mathcal{N}_P \cup \mathcal{N}_N\}}\exp(\text{sim}(\tilde{\mathbf{p}}_i, \tilde{\mathbf{p}}_j))}.
\end{equation}
We sum $\mathcal{L}_i$ all nodes to obtain the final regulation loss, i.e., $\mathcal{L}_\text{reg} = \sum_{N} \mathcal{L}_i$.
The detailed training procedure can be found in Algorithm~\ref{alg:rgvq}. The proposed regularization term is added to the reconstruction loss in Equation~\ref{eq: vqloss}, forming the ultimate loss:
\begin{equation}
{     \mathcal{L}_{\text{VQ}} = 
\mathcal{L}_{\text{recon}}+ 
    \|\text{sg}[\mathbf{H}] - \mathbf{ \tilde Z}\|^2 + 
    \beta \|\mathbf{H} - \text{sg}[\mathbf{ \tilde Z}]\|^2 + \mathcal{L}_\text{reg}.}
\label{eq: allloss}
\end{equation}

\section{Experiments}
We evaluate the performance of RGVQ in terms (1) codebook utilization, (2) transferability, and (3) serialization.
First, we evaluate the codebook utilization to verify its ability to mitigate codebook collapse.
Second, we investigate the transferability.
We integrate RGVQ into GFT~\cite{wang2024gft}, a graph foundation model that utilizes the learned codebook as pretrained graph tokens, and evaluate the performance on cross-task and cross-domain graphs.
Third, we assess the serialization capability of RGVQ by examining its compatibility with sequence-based models. 
We use GQT~\cite{wang2025learning}, a transformer taking VQ tokens as input sequences, and evaluate the performance on node classification. 
Detailed dataset statistics, baselines, implementation details are provided in Appendix~\ref{appendix:experiments}. 
Experiments on large graphs can also be found in Appendix~\ref{appendix:complexity}.
\begin{table*}[t]
\centering
\caption{Codebook utilization on homophilous and heterophilous graphs with codebook size $K = 512$. \textbf{Bold} highlights the best performance.}
\resizebox{\textwidth}{!}{
\begin{tabular}{l|cccccc|ccc}
\toprule
& Cora & PubMed & Citeseer & Photo & Computer & WikiCS & Ratings & Roman & Questions \\
\midrule
Graph VQ     & \makecell{94.47$\pm$8.65} & \makecell{4.14$\pm$1.03} & \makecell{60.09$\pm$5.59} & \makecell{1.00$\pm$0.00} & \makecell{1.00$\pm$0.00} & \makecell{10.18$\pm$2.13} & \makecell{13.29$\pm$2.89} & \makecell{10.84$\pm$3.48} & \makecell{20.78$\pm$3.65} \\
EMA          & \makecell{91.68$\pm$9.17} & \makecell{5.12$\pm$1.46} & \makecell{55.15$\pm$6.73} & \makecell{1.00$\pm$0.00} & \makecell{1.00$\pm$0.00} & \makecell{11.27$\pm$3.36} & \makecell{9.12$\pm$2.33} & \makecell{6.20$\pm$3.24} & \makecell{14.15$\pm$3.51} \\
AP           & \makecell{75.32$\pm$6.28} & \makecell{126.55$\pm$12.64} & \makecell{9.03$\pm$2.16} & \makecell{54.95$\pm$5.43} & \makecell{59.33$\pm$8.55} & \makecell{83.55$\pm$9.86} & \makecell{73.82$\pm$8.14} & \makecell{118.46$\pm$16.21} & \makecell{66.57$\pm$8.27} \\
Reset        & \makecell{65.79$\pm$8.56} & \makecell{102.78$\pm$15.78} & \makecell{85.19$\pm$4.41} & \makecell{10.73$\pm$1.98} & \makecell{17.18$\pm$2.37} & \makecell{134.44$\pm$7.35} & \makecell{130.83$\pm$8.88} & \makecell{150.51$\pm$11.15} & \makecell{141.98$\pm$10.11} \\
PT           & \makecell{60.57$\pm$10.25} & \makecell{6.17$\pm$1.12} & \makecell{138.98$\pm$10.54} & \makecell{3.78$\pm$1.37} & \makecell{2.94$\pm$1.27} & \makecell{3.10$\pm$1.31} & \makecell{37.65$\pm$5.76} & \makecell{14.49$\pm$2.52} & \makecell{58.99$\pm$8.34} \\
SimVQ        & \makecell{40.09$\pm$6.53} & \makecell{23.96$\pm$2.56} & \makecell{38.11$\pm$6.67} & \makecell{37.29$\pm$4.85} & \makecell{40.47$\pm$6.54} & \makecell{45.90$\pm$7.35} & \makecell{16.08$\pm$4.11} & \makecell{42.22$\pm$8.34} & \makecell{21.71$\pm$5.27} \\
HQA-GAE      & \makecell{130.06$\pm$5.52} & \makecell{164.77$\pm$14.15} & \makecell{93.67$\pm$11.32} & \makecell{166.32$\pm$10.98} & \makecell{114.08$\pm$10.15} & \makecell{98.73$\pm$7.82} & \makecell{92.17$\pm$8.66} & \makecell{89.05$\pm$8.23} & \makecell{72.86$\pm$7.79} \\
\midrule
RGVQ         & \makecell{\textbf{211.69}$\pm$5.27} & \makecell{\textbf{319.09}$\pm$10.40} & \makecell{\textbf{188.17}$\pm$11.23} & \makecell{\textbf{446.02}$\pm$15.82} & \makecell{\textbf{413.10}$\pm$10.78} & \makecell{\textbf{228.82}$\pm$5.96} & \makecell{\textbf{200.93}$\pm$7.89} & \makecell{\textbf{374.51}$\pm$11.13} & \makecell{\textbf{250.79}$\pm$8.63} \\
\bottomrule
\end{tabular}
}
\label{tab:performance1}
\end{table*}
\begin{table*}[t]
\centering
\caption{Cross-domain and cross-task performance in the pre-training and fine-tuning setting. { Metrics are reported in terms of ROC-AUC for Graph Classification and Accuracy for all other tasks.} \textbf{Bold} highlight the best performance.}
\resizebox{\textwidth}{!}{
\begin{tabular}{l|ccc|cc|cc|c}
\toprule
\multirow{2}{*}{Method} 
  & \multicolumn{3}{c|}{\textbf{Node Classification}} 
  & \multicolumn{2}{c|}{\textbf{Link Classification}} 
  & \multicolumn{2}{c}{\textbf{Graph Classification}} & \\
  & Cora & PubMed & WikiCS & WN18RR & FB15K237 & HIV & PCBA &\textit{Avg.} \\
\midrule
GCN             & \makecell{75.65$\pm$1.37} & \makecell{75.61$\pm$2.10} & \makecell{75.28$\pm$1.34} & \makecell{73.79$\pm$0.39} & \makecell{82.22$\pm$0.28} & \makecell{64.84$\pm$4.78} & \makecell{71.32$\pm$0.49} & \makecell{74.10} \\
GAT             & \makecell{76.24$\pm$1.62} & \makecell{74.86$\pm$1.87} & \makecell{76.28$\pm$0.78} & \makecell{80.16$\pm$0.27} & \makecell{88.93$\pm$0.15} & \makecell{65.54$\pm$6.93} & \makecell{70.12$\pm$0.89} & \makecell{76.01} \\
GIN             & \makecell{73.59$\pm$2.10} & \makecell{69.51$\pm$6.87} & \makecell{49.77$\pm$4.72} & \makecell{74.02$\pm$0.55} & \makecell{83.21$\pm$0.53} & \makecell{66.86$\pm$3.48} & \makecell{72.69$\pm$0.22} & \makecell{69.95} \\
\midrule
DGI             & \makecell{72.10$\pm$0.34} & \makecell{73.13$\pm$0.64} & \makecell{75.32$\pm$0.95} & \makecell{75.75$\pm$0.59} & \makecell{81.34$\pm$0.15} & \makecell{59.62$\pm$1.21} & \makecell{63.31$\pm$0.89} & \makecell{71.51} \\
BGRL            & \makecell{71.20$\pm$0.30} & \makecell{75.29$\pm$1.33} & \makecell{76.53$\pm$0.69} & \makecell{75.44$\pm$0.30} & \makecell{80.66$\pm$0.29} & \makecell{63.95$\pm$1.06} & \makecell{67.09$\pm$1.00} & \makecell{72.88} \\
GraphMAE        & \makecell{73.10$\pm$0.40} & \makecell{74.32$\pm$0.33} & \makecell{72.61$\pm$0.39} & \makecell{78.99$\pm$0.48} & \makecell{85.30$\pm$0.16} & \makecell{61.04$\pm$0.55} & \makecell{63.30$\pm$0.78} & \makecell{72.66} \\
GIANT        & \makecell{75.13$\pm$0.49} & \makecell{72.31$\pm$0.53} & \makecell{76.56$\pm$0.88} & \makecell{84.36$\pm$0.30} & \makecell{87.45$\pm$0.54} & \makecell{65.44$\pm$1.39} & \makecell{61.49$\pm$0.99} & \makecell{74.68} \\
\midrule
GFT             & \makecell{78.35$\pm$1.07} & \makecell{73.39$\pm$1.68} & \makecell{79.13$\pm$0.32} & \makecell{90.87$\pm$0.25} & \makecell{89.89$\pm$0.27} & \makecell{72.16$\pm$1.69} & \makecell{72.74$\pm$1.23} & \makecell{79.50} \\
{GFT + EMA}            
    & \makecell{{79.44$\pm$0.89}}
    & \makecell{{74.01$\pm$1.57}}
    & \makecell{{78.94$\pm$0.41}}
    & \makecell{{90.58$\pm$0.43}}
    & \makecell{{89.75$\pm$0.19}}
    & \makecell{{72.39$\pm$1.52}}
    & \makecell{{73.04$\pm$1.01}}
    & \makecell{{79.73}} \\
{GFT + AP}            
    & \makecell{{79.69$\pm$1.07}}
    & \makecell{{75.05$\pm$0.86}}
    & \makecell{{79.73$\pm$0.35}}
    & \makecell{{89.56$\pm$0.18}}
    & \makecell{{89.05$\pm$0.18}}
    & \makecell{{71.86$\pm$1.53}}
    & \makecell{{71.48$\pm$0.99}}
    & \makecell{{79.48}} \\
{GFT + Reset}            
    & \makecell{{80.07$\pm$0.91}}
    & \makecell{{75.51$\pm$0.69}}
    & \makecell{{79.85$\pm$0.33}}
    & \makecell{{91.18$\pm$0.43}}
    & \makecell{{88.09$\pm$0.23}}
    & \makecell{{72.79$\pm$1.65}}
    & \makecell{{71.95$\pm$0.85}}
    & \makecell{{79.92}} \\
{GFT + PT}            
    & \makecell{{78.57$\pm$0.86}}
    & \makecell{{74.12$\pm$1.05}}
    & \makecell{{72.75$\pm$1.72}}
    & \makecell{{88.63$\pm$0.15}}
    & \makecell{{88.45$\pm$0.17}}
    & \makecell{{71.01$\pm$1.74}}
    & \makecell{{73.73$\pm$1.12}}
    & \makecell{{78.18}} \\
{GFT + SimVQ}             
    & \makecell{{77.61$\pm$0.73}}
    & \makecell{{76.41$\pm$1.28}}
    & \makecell{{76.57$\pm$0.68}}
    & \makecell{{82.72$\pm$0.53}}
    & \makecell{{82.03$\pm$0.35}}
    & \makecell{{66.57$\pm$1.35}}
    & \makecell{{69.90$\pm$0.91}}
    & \makecell{{75.97}} \\

GFT + RGVQ      & \makecell{\textbf{80.85}$\pm$0.73} & \makecell{\textbf{77.46$\pm$0.94}} & \makecell{\textbf{80.10$\pm$0.52}} & \makecell{\textbf{91.32$\pm$0.26}} & \makecell{\textbf{90.45$\pm$0.31}} & \makecell{\textbf{74.10$\pm$1.49}} & \makecell{\textbf{75.68$\pm$0.99}} & \makecell{\textbf{81.42}} \\
\bottomrule
\end{tabular}
}
\label{tab:performance2}
\end{table*}

\begin{table*}[t]
\centering
\caption{Mean performance on node classification tasks. Metrics are reported in terms of ROC-AUC for Questions, and Accuracy for all other datasets. \textbf{Bold} indicates the best performance.}
\resizebox{0.98\textwidth}{!}{
\begin{tabular}{l|cccccc|ccc}
\toprule
& Cora & PubMed & Citeseer & Photo & Computer & WikiCS & Ratings & Roman & Questions \\
\midrule
GCN         & \makecell{75.65$\pm$1.37} & \makecell{78.80$\pm$0.60} & \makecell{71.60$\pm$0.40} & \makecell{92.70$\pm$0.20} & \makecell{89.65$\pm$0.52} & \makecell{77.47$\pm$0.85} & \makecell{48.70$\pm$0.63} & \makecell{73.69$\pm$0.74} & \makecell{76.09$\pm$1.27} \\
GAT         & \makecell{76.24$\pm$1.62} & \makecell{79.00$\pm$0.40} & \makecell{72.10$\pm$1.10} & \makecell{93.87$\pm$0.11} & \makecell{90.78$\pm$0.13} & \makecell{76.91$\pm$0.82} & \makecell{52.70$\pm$0.62} & \makecell{88.75$\pm$0.41} & \makecell{76.79$\pm$0.71} \\
\midrule
GraphGPS    & \makecell{82.84$\pm$1.03} & \makecell{79.94$\pm$0.26} & \makecell{72.73$\pm$1.23} & \makecell{95.06$\pm$0.13} & \makecell{91.19$\pm$0.54} & \makecell{78.66$\pm$0.49} & \makecell{53.10$\pm$0.42} & \makecell{82.00$\pm$0.61} & \makecell{71.73$\pm$1.47} \\
SGFormer    & \makecell{84.50$\pm$0.80} & \makecell{80.30$\pm$0.60} & \makecell{72.60$\pm$0.20} & \makecell{95.10$\pm$0.47} & \makecell{91.99$\pm$0.76} & \makecell{73.46$\pm$0.56} & \makecell{48.01$\pm$0.49} & \makecell{79.10$\pm$0.32} & \makecell{72.15$\pm$1.31} \\
Exphomer    & \makecell{82.77$\pm$1.38} & \makecell{79.46$\pm$0.35} & \makecell{71.63$\pm$1.19} & \makecell{95.35$\pm$0.22} & \makecell{91.47$\pm$0.17} & \makecell{78.54$\pm$0.49} & \makecell{53.51$\pm$0.46} & \makecell{89.03$\pm$0.37} & \makecell{-} \\
NodeFormer    & \makecell{83.20$\pm$0.90} & \makecell{79.90$\pm$1.00} & \makecell{72.50$\pm$1.10} & \makecell{93.46$\pm$0.35} & \makecell{86.98$\pm$0.62} & \makecell{74.73$\pm$0.94} & \makecell{43.86$\pm$0.35} & \makecell{64.49$\pm$0.73} & \makecell{74.27$\pm$1.46} \\
\midrule
GQT         & \makecell{86.44$\pm$1.58} & \makecell{81.60$\pm$1.35} & \makecell{73.14$\pm$1.26} & \makecell{94.46$\pm$0.68} & \makecell{92.13$\pm$0.23} & \makecell{80.03$\pm$0.19} & \makecell{54.04$\pm$0.12} & \makecell{89.85$\pm$0.73} & \makecell{76.52$\pm$1.52} \\
{GQT + EMA}         & {{86.23$\pm$1.19}} & {\makecell{81.41$\pm$1.24}} & {\makecell{73.08$\pm$1.58}} & {\makecell{94.01$\pm$0.57}} & {\makecell{91.95$\pm$0.18}} & {\makecell{79.98$\pm$0.23}} & {\makecell{54.10$\pm$0.08}} & {\makecell{89.91$\pm$0.51}} & {\makecell{75.94$\pm$1.16}} \\
{GQT + AP}        & \makecell{{85.89$\pm$0.94}} & \makecell{{83.31$\pm$0.97}} & \makecell{{72.56$\pm$1.38}} & \makecell{{96.15$\pm$1.21}} & \makecell{{94.46$\pm$0.36}} & \makecell{{82.03$\pm$0.59}} & \makecell{{54.54$\pm$0.24}} & \makecell{{90.46$\pm$0.52}} & \makecell{{76.96$\pm$1.17}} \\
{GQT + Reset}       & \makecell{{86.15$\pm$1.07}} & \makecell{{83.50$\pm$1.01}} & \makecell{{71.59$\pm$1.37}} & \makecell{{95.15$\pm$0.55}} & \makecell{{94.79$\pm$0.48}} & \makecell{{82.84$\pm$0.23}} & \makecell{{54.41$\pm$0.17}} & \makecell{{90.50$\pm$0.42}} & \makecell{{78.13$\pm$0.98}} \\
{GQT + PT}        & \makecell{{85.71$\pm$1.44}} & \makecell{{80.92$\pm$1.15}} & \makecell{{79.53$\pm$1.23}} & \makecell{{94.74$\pm$0.76}} & \makecell{{92.35$\pm$0.35}} & \makecell{{75.65$\pm$0.78}} & \makecell{{54.50$\pm$0.14}} & \makecell{{89.76$\pm$0.68}} & \makecell{{76.74$\pm$1.34}} \\
{GQT + SimVQ}        & \makecell{{86.02$\pm$1.64}} & \makecell{{82.56$\pm$1.02}} & \makecell{{72.58$\pm$1.14}} & \makecell{{95.21$\pm$0.77}} & \makecell{{94.23$\pm$0.21}} & \makecell{{81.78$\pm$0.32}} & \makecell{{53.98$\pm$0.15}} & \makecell{{90.15$\pm$0.66}} & \makecell{{76.35$\pm$1.21}} \\
GQT + RGVQ & \makecell{\textbf{88.34}$\pm$1.32} & \makecell{\textbf{86.54$\pm$1.41}} & \makecell{\textbf{81.25$\pm$1.01}} & \makecell{\textbf{97.66$\pm$1.05}} & \makecell{\textbf{95.67$\pm$0.36}} & \makecell{\textbf{83.58$\pm$0.66}} & \makecell{\textbf{55.16$\pm$0.19}} & \makecell{\textbf{90.98$\pm$0.66}} & \makecell{\textbf{78.26$\pm$1.07}}
 \\
\bottomrule
\end{tabular}
}
\label{tab:performance3}
\end{table*}



\vspace{-1mm}

\subsection{Performance Evaluation}
\myparagraph{Codebook Utilization}
We evaluate codebook utilization by comparing vanilla Graph VQ and its variants: EMA~\cite{lancucki2020ema}, affine parameters (AP)~\cite{huh2023straightening}, codebook reset (Reset)~\cite{zeghidour2021reset}, and pretrained encoders (PT)~\cite{zhao2024pretrainVQ}, as well as existing SOTA VQ models: SimVQ~\cite{zhu2024softassign} and HQA-GAE~\cite{zeng2025hgvq}. 
All methods use orthogonal normalization, cosine similarity, and K-Means initialization~\cite{im2023vector}, and are trained on the feature and link reconstruction tasks. We fix the codebook size at 512 and report perplexity in Table~\ref{tab:performance1}. 
Across all datasets, RGVQ outperforms all baselines by a clear margin and remains robust, showing that Gumbel-Softmax reparameterization combined with structure-aware regularization leads to more balanced codebook use, prevents collapse, and enables more expressive graph representations.
By contrast, vanilla Graph VQ and its variants suffer from severe codebook collapse, with perplexity values as low as 1.00 in several datasets. 
More advanced mitigation strategies like SimVQ and HQA-GAE offer only small improvements.

\myparagraph{Transferability}
To evaluate the effectiveness of RGVQ in learning transferrable graph tokens, we integrate it into a graph foundation model, i.e., GFT, and compare it with vanilla Graph VQ and its variants with SOTA mitigation strategies.
Moreover, we include supervised GNNs, i.e., GCN~\cite{zhang2022semi}, GAT~\cite{velivckovic2017graph}, and GIN~\cite{xu2018powerful}, and graph self-supervised methods, i.e., DGI~\cite{velivckovic2018dgi}, BGRL~\cite{thakoor2021bgrl}, GraphMAE~\cite{hou2022graphmae}, and GIANT~\cite{chien2021node}.
The supervised GNNs are trained directly on target dataset, while the self-supervised methods and all GFT variants are pretrained on the full set of datasets and then fine-tuned per target.
Table~\ref{tab:performance2} reports cross-domain and cross-task performance.
RGVQ consistently achieves the best performance across all tasks and datasets, surpassing both supervised and self-supervised models.
It not only ensures better codebook utilization but also yields consistent downstream improvements. 

\myparagraph{Serialization}
To further evaluate the effectiveness of RGVQ in serialization, we integrate it into the Graph Quantized Transformer (GQT), where discrete tokens serve as the input sequence to a vanilla Transformer backbone. 
We follow the original sequence reconstruction method~\cite{wang2025learning} and compare the performance of RGVQ-enhanced GQT against GQT with different anti-collapse methods, supervised GNNs, and graph transformers, including GraphGPS~\cite{rampavsek2022graphgps}, SGFormer~\cite{wu2023sgformer}, Exphomer~\cite{shirzad2023exphormer}, and NodeFormer\citep{wu2022nodeformer}.
Table~\ref{tab:performance3} summarizes the node classification results across various benchmarks.  Compared to GQT with conventional anti-collapse methods, incorporating RGVQ consistently improves classification accuracy on most datasets.
While gains on some datasets are modest, this is expected, as discrete tokens serve as intermediate representations.
In this setting, the backbone architecture (e.g., GQT’s Residual VQ) and supervision during fine-tuning largely determine the final performance and can compensate for imperfections in tokenization.

\subsection{Ablation Studies}\label{Exp:abl}
We further conduct ablation experiments on the contributions of the Gumbel–Softmax reparameterization and the structure-aware contrastive regularization. Then, we vary the key hyperparameters in each module to evaluate their individual contributions. In addition, we provide additional ablation studies, convergence analysis,
and an investigation of the effect of the number of GNN layers in
Appendix~\ref{appendix:additional}.

\myparagraph{Drop-one Ablation}
\begin{table}[t]
\centering
\small
\caption{Drop-one ablation.}
\setlength{\tabcolsep}{3pt}
\begin{tabular}{l|cc|cc|cc}
\toprule
\multirow{2}{*}{Variant}
  & \multicolumn{2}{c|}{\textbf{Cora}}
  & \multicolumn{2}{c|}{\textbf{PubMed}}
  & \multicolumn{2}{c}{\textbf{WikiCS}} \\
  & Perp. & Acc. & Perp. & Acc. & Perp. & Acc. \\
\midrule
Variant-1& 94.47  & 78.35 & 4.14   & 73.39 & 10.18  & 79.13 \\
Variant-2 &  172.32 & 79.87 & 215.35 & 76.32 & 153.35 & 79.84 \\
Variant-3 & 135.45 & 79.12 & 208.16 & 76.29 & 179.49 & 79.79 \\
RGVQ      & 
           \textbf{211.69} & \textbf{80.85}
           & \textbf{319.09} & \textbf{77.46}
           & \textbf{228.82} & \textbf{80.10} \\
\bottomrule
\end{tabular}
\label{tab:ablation}
\end{table}
As the structure-aware regularization relies on the soft assignment probabilities produced by the Gumbel–Softmax reparameterization, without reparameterization, one-hot assignments provide no gradient to inactive codewords, making the regularization ineffective. 
Conversely, using soft assignments without regularization does not constrain the assignment distribution and therefore makes RGVQ behave similarly to vanilla VQ. 
Therefore, we remove either structural signals or feature signals to construct contrastive sets and assess three variants: removing all proposed components (variant-1), reparameterization with only structural regularization (variant-2), and reparameterization with only feature regularization (variant-3).
As shown in Table~\ref{tab:ablation}, excluding either structural signals or feature signals reduces quantization diversity and consequently harms downstream accuracy. 
Removing Gumbel-softmax causes RGVQ to degenerate to vanilla VQ and leads to severe collapse.
These results indicate that Gumbel-Softmax reparameterization and structure-aware regularization are mutually dependent in preventing codebook collapse, and both topological and feature information are essential for enhancing quantization diversity.

\begin{figure*}[t]
  \centering
  \begin{subfigure}[t]{0.32\linewidth}
    \centering
    \includegraphics[width=0.95\linewidth]{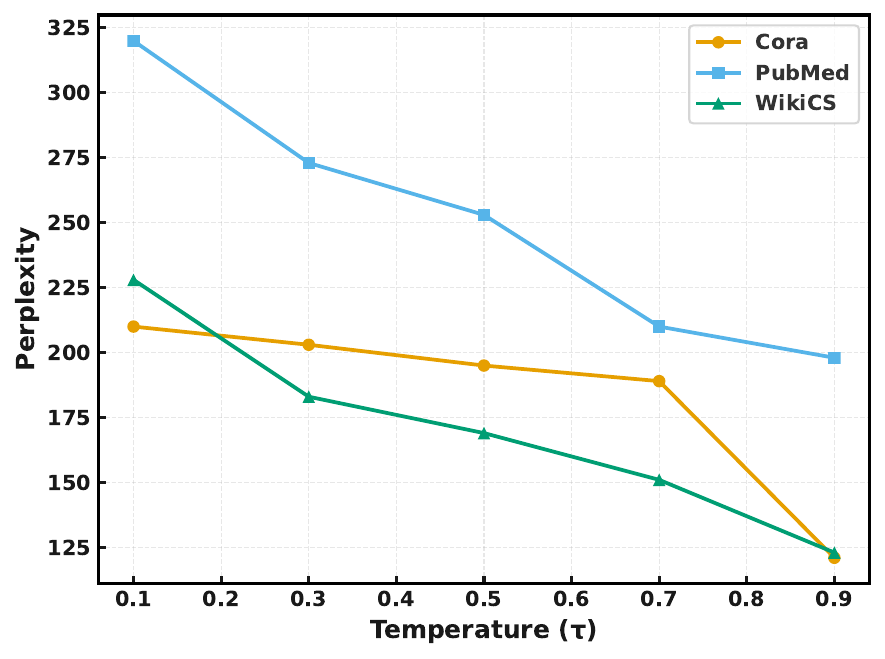}
    \caption{Temperature $(\tau)$.}
    \label{fig:temp}
  \end{subfigure}
  \hfill
  \begin{subfigure}[t]{0.32\linewidth}
    \centering
    \includegraphics[width=0.95\linewidth]{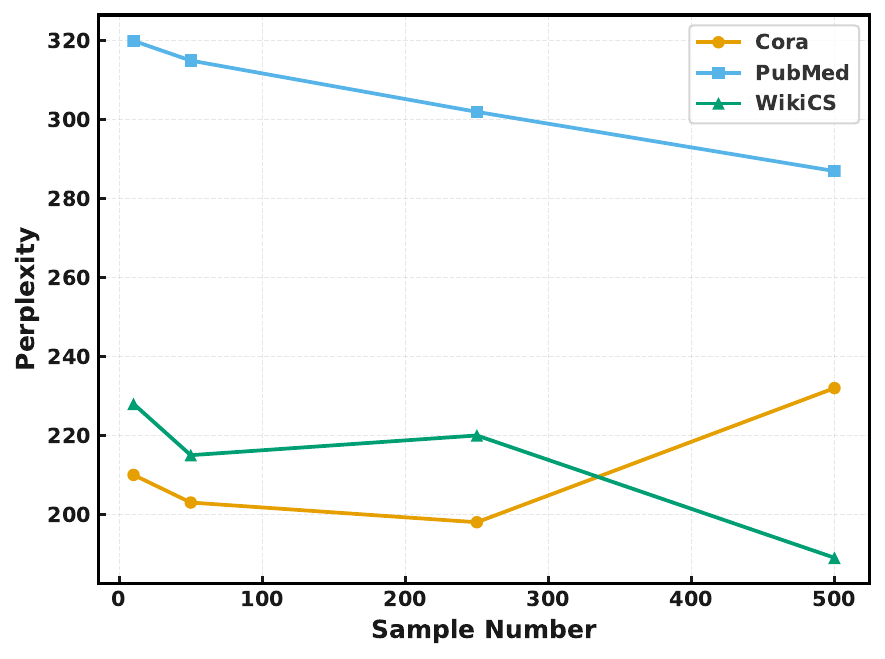}
    \caption{Contrastive Samples.}
    \label{fig:sample}
  \end{subfigure}
  \hfill
  \begin{subfigure}[t]{0.32\linewidth}
    \centering
    \includegraphics[width=0.95\linewidth]{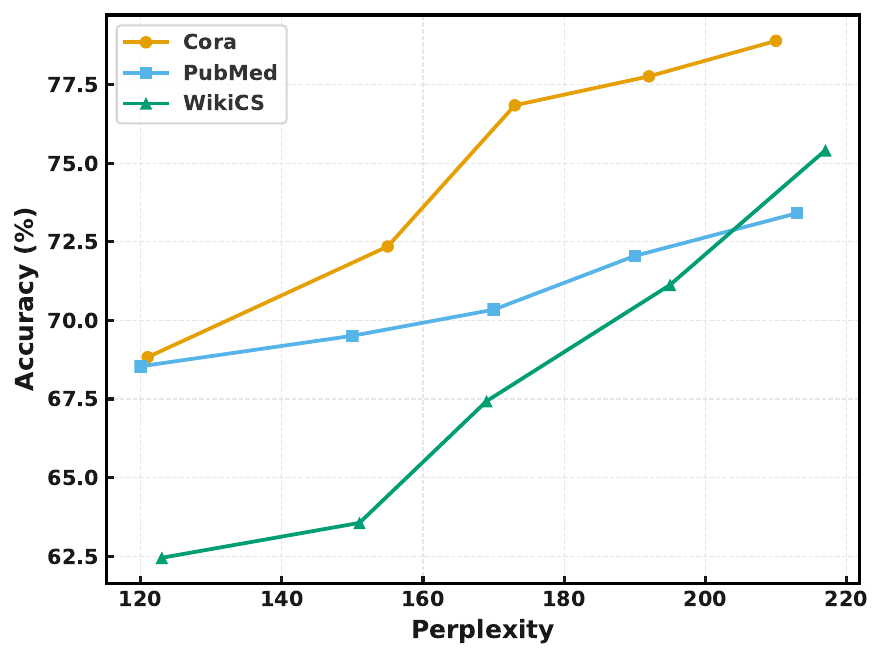}
    \caption{Downstream Performance.}
    \label{fig:property_vs_perplexity}
  \end{subfigure}
  \caption{Ablation study with varying parameters.}
  \label{fig:ablation}
\end{figure*}

\myparagraph{Influence of the Temperature}
We investigate how the Gumbel–Softmax temperature $\tau$ affects perplexity. 
As shown in Figure~\ref{fig:temp}, lower temperatures, which produce distributions closer to one-hot, consistently improve codebook utilization.
This indicates that, unlike some prior work~\cite{zeng2025hgvq} that rely on temperature annealing, a relatively low and fixed temperature is sufficient to regularize the codebook distributions and address the non-differentiability of deterministic VQ.


\begin{table*}[t]
\centering
\caption{Ablation performance of variants under varying levels of heterophily.}
\resizebox{\textwidth}{!}{
\begin{tabular}{lcccccccccccc}
\toprule
\multirow{2}{*}{Variant}
& \multicolumn{4}{c}{PubMed (0.19)}
& \multicolumn{4}{c}{Ratings (0.61)}
& \multicolumn{4}{c}{Roman (0.95)} \\
\cmidrule(lr){2-5}
\cmidrule(lr){6-9}
\cmidrule(lr){10-13}
& Perp.
& Acc.
& Feat. Recon.
& Link Recon.
& Perp.
& Acc.
& Feat. Recon.
& Link Recon.
& Perp.
& Acc.
& Feat. Recon.
& Link Recon. \\
\midrule

Feat. only
& 290.45
& 75.85
& \textbf{3.45e-4}
& 9.14e-1
& 194.41
& 51.91
& \textbf{2.07e-3}
& 8.90e-1
& \textbf{379.33}
& \textbf{88.20}
& \textbf{3.54e-3}
& 9.06e-1 \\

Adj. only
& 245.46
& 76.42
& 3.06e-3
& \textbf{6.59e-1}
& 163.31
& 51.72
& 5.98e-3
& \textbf{6.69e-1}
& 284.48
& 83.62
& 1.27e-2
& \textbf{6.80e-1} \\

Feat. \& Adj.
& 154.14
& 74.13
& 3.79e-4
& 7.31e-1
& 89.96
& 47.08
& 2.91e-3
& 7.73e-1
& 181.60
& 83.08
& 4.83e-3
& 7.54e-1 \\

\textbf{RGVQ}
& \textbf{319.09}
& \textbf{77.46}
& 3.52e-4
& 6.93e-1
& \textbf{200.93}
& \textbf{52.83}
& 2.10e-3
& 6.93e-1
& 374.51
& 87.49
& 3.65e-3
& 6.93e-1 \\

\bottomrule
\end{tabular}
}
\label{tab:ablation_hetero}
\end{table*}

\myparagraph{Influence of Contrastive Samples}
We examine how the number of contrastive samples in structure-aware regularization affects perplexity.
As shown in Figure~\ref{fig:sample}, even a small number of contrastive samples (e.g., $n=10$ or $50$) achieves relatively high perplexity, indicating effective codebook utilization.
Increasing $n$ does not consistently lead to better utilization and may even degrade in some datasets, potentially due to added training noise.

\myparagraph{Perplexity and Downstream Performance}
To understand the relations between token diversity and the downstream performance, we evaluate how perplexity affects the node classification results. We select different pretraining checkpoints of RGVQ to reflect different perplexities.
At the downstream stage, we utilize the pretrained tokens and finetune with node labels.
Based on Figure~\ref{fig:property_vs_perplexity}, we observe a consistent positive correlation between perplexity and accuracy.
This suggests that a more diverse codebook can capture finer-grained structural patterns, enabling the model to learn more discriminative embeddings. It should be noted that while the optimal codebook size may vary depending on the trade-off between compression and expressiveness, our model provides a flexible framework that maintains token diversity and scales to larger codebook sizes.

\myparagraph{Influence of Heterophily}
To evaluate the effectiveness of contrastive set design under varying levels of heterophily, we compare four variants for positive sets: feature-similar samples only (Feat. only), adjacent-only (Adj. only), their intersection (Feat. \& Adj.), and RGVQ, while keeping the negative sets unchanged as defined in Equation~\ref{eq:negative}. Table~\ref{tab:ablation_hetero} summarizes quantization results, node classification and graph reconstruction performance on graph datasets with varying heterophily ratios.
First, with respect to perplexity and downstream performance, feature-only positive sets achieve comparable perplexity to RGVQ, as they provide sufficient positives to regularize assignments and prevent collapse. However, lacking structural constraints, neighboring nodes may be quantized into different tokens, which can impair downstream performance in graphs with homophily. In highly heterophilous graphs, such as Roman-empire, feature-only positives slightly outperform RGVQ, since adjacent-only positives may introduce semantic noise.
For adjacent-only and intersection-based variants, the absence of feature-level constraints weakens semantic consistency, and the limited number of positive samples causes InfoNCE to be dominated by negatives, leading to unstable optimization and degraded performance.
Regarding reconstruction performance, RGVQ achieves the best balance between feature and link reconstruction. Omitting either feature-similar or adjacency positives results in a clear trade-off, indicating that single-type positives cannot fully preserve graph information.

\section{Conclusion}
In this paper, we investigate the codebook collapse in Graph VQ.
Through empirical studies, we show that codebook collapse is not incidental, but a systematic issue in graphs.
We diagnose underlying causes and propose RGVQ, a differentiable method that integrates both graph topology and feature similarity as explicit regularization to enhance codebook utilization.
Extensive experiments demonstrate that RGVQ significantly mitigates collapse and improves downstream performance, highlighting its applicability in learning expressive and transferable graph representations.

\section*{Impact Statement}
\myparagraph{Ethical Aspects}
To the best of our knowledge, this work does not raise any specific ethical concerns.
All datasets used in our experiments are publicly available benchmark datasets and
have been widely adopted in prior research. Our method focuses on improving
representation learning techniques and does not involve sensitive attributes.

\myparagraph{Societal Consequences}
This work studies a fundamental methodological issue in graph machine learning,
namely codebook collapse in discrete graph tokenization, with the goal of improving
model robustness and representation quality.
The contributions of this paper are empirical studies of vector quantization on graphs and previous methods, the diagnosis of codebook collapse, and the proposed solution.
We do not identify any immediate or specific societal consequences
that warrant special discussion here.

\section*{Acknowledgements}
Xiaoyang Wang is supported by ARC DP240101322 and DP260100689. Wenjie Zhang is supported by Australian Research Council Centre of Excellence for Mathematical Modelling of Cellular Systems CE230100001 and Australian Research Council Discovery Project DP260100689.

\bibliography{example_paper}

@article{wang2024gft,
  title={Gft: Graph foundation model with transferable tree vocabulary},
  author={Wang, Zehong and Zhang, Zheyuan and Chawla, Nitesh and Zhang, Chuxu and Ye, Yanfang},
  journal={Advances in Neural Information Processing Systems},
  volume={37},
  pages={107403--107443},
  year={2024}
}

@inproceedings{zhai2025sgpt,
  title={SGPT: Few-Shot Prompt Tuning for Signed Graphs},
  author={Zhai, Zian and Sima, Qing and Wang, Xiaoyang and Zhang, Wenjie},
  booktitle={Proceedings of the 34th ACM International Conference on Information and Knowledge Management},
  pages={4045--4055},
  year={2025}
}

@inproceedings{
luo2024nodeid,
title={Node Identifiers: Compact, Discrete Representations for Efficient Graph Learning},
author={Yuankai Luo and Hongkang Li and Qijiong Liu and Lei Shi and Xiao-Ming Wu},
booktitle={The Thirteenth International Conference on Learning Representations},
year={2025}
}

@inproceedings{wang2025learning,
  title={Learning graph quantized tokenizers},
  author={Wang, Limei and Hassani, Kaveh and Zhang, Si and Fu, Dongqi and Yuan, Baichuan and Cong, Weilin and Hua, Zhigang and Wu, Hao and Yao, Ning and Long, Bo},
  booktitle={The Thirteenth International Conference on Learning Representations},
  year={2025}
}

@article{li2024autoregressive,
  title={Autoregressive image generation without vector quantization},
  author={Li, Tianhong and Tian, Yonglong and Li, He and Deng, Mingyang and He, Kaiming},
  journal={Advances in Neural Information Processing Systems},
  volume={37},
  pages={56424--56445},
  year={2024}
}

@inproceedings{zhang2024codebook,
  title={Codebook transfer with part-of-speech for vector-quantized image modeling},
  author={Zhang, Baoquan and Wang, Huaibin and Luo, Chuyao and Li, Xutao and Liang, Guotao and Ye, Yunming and Qi, Xiaochen and He, Yao},
  booktitle={Proceedings of the IEEE/CVF Conference on Computer Vision and Pattern Recognition},
  pages={7757--7766},
  year={2024}
}

@inproceedings{zhang2023VQimage,
  title={Regularized vector quantization for tokenized image synthesis},
  author={Zhang, Jiahui and Zhan, Fangneng and Theobalt, Christian and Lu, Shijian},
  booktitle={Proceedings of the IEEE/CVF Conference on Computer Vision and Pattern Recognition},
  pages={18467--18476},
  year={2023}
}

@inproceedings{zhu2024softassign,
  title={Addressing representation collapse in vector quantized models with one linear layer},
  author={Zhu, Yongxin and Li, Bocheng and Xin, Yifei and Xia, Zhihua and Xu, Linli},
  booktitle={Proceedings of the IEEE/CVF International Conference on Computer Vision},
  pages={22968--22977},
  year={2025}
}

@article{zhao2024pretrainVQ,
  title={Representation Collapsing Problems in Vector Quantization},
  author={Zhao, Wenhao and Zou, Qiran and Shah, Rushi and Liu, Dianbo},
  journal={arXiv preprint arXiv:2411.16550},
  year={2024}
}

@inproceedings{
li2025dhg,
title={{DHG}-Bench: A Comprehensive Benchmark for Deep Hypergraph Learning},
author={Fan Li and Xiaoyang Wang and Wenjie Zhang and Ying Zhang and Xuemin Lin},
booktitle={The Fourteenth International Conference on Learning Representations},
year={2026}
}

@inproceedings{tan2026memotime,
  title={Memotime: Memory-augmented temporal knowledge graph enhanced large language model reasoning},
  author={Tan, Xingyu and Wang, Xiaoyang and Liu, Qing and Xu, Xiwei and Yuan, Xin and Zhu, Liming and Zhang, Wenjie},
  booktitle={Proceedings of the ACM Web Conference 2026},
  pages={4220--4231},
  year={2026}
}

@article{tan2026privgemo,
  title={PrivGemo: Privacy-Preserving Dual-Tower Graph Retrieval for Empowering LLM Reasoning with Memory Augmentation},
  author={Tan, Xingyu and Wang, Xiaoyang and Liu, Qing and Xu, Xiwei and Yuan, Xin and Zhu, Liming and Zhang, Wenjie},
  journal={arXiv preprint arXiv:2601.08739},
  year={2026}
}

@article{zheng2026rethinkinggatingmechanismsparse,
  title={Rethinking gating mechanism in sparse MOE: handling arbitrary modality inputs with confidence-guided gate},
  author={Zheng, Liangwei Nathan and Zhang, Wei Emma and Guo, Mingyu and Xu, Miao and Maennel, Olaf and Chen, Weitong},
  journal={arXiv preprint arXiv:2505.19525},
  year={2025}
}

@article{xu2026c2tctrainingfreeframeworkefficient,
  title={C2{TC}: A Training-Free Framework for Efficient Tabular Data Condensation},
  author={Xu, Sijia and Li, Fan and Wang, Xiaoyang and Yang, Zhengyi and Lin, Xuemin},
  journal={arXiv preprint arXiv:2602.21717},
  year={2026}
}

@article{li2026fairness,
  title={Fairness-aware Hypergraph Self-Supervised Learning with Sampling-efficient Signals},
  author={Li, Fan and Wang, Xiaoyang and Cheng, Dawei and Zhang, Ying and Zhang, Wenjie and Lin, Xuemin},
  journal={IEEE Transactions on Knowledge and Data Engineering},
  year={2026},
  publisher={IEEE}
}

@inproceedings{
yang2023vqgraph,
title={{VQG}raph: Rethinking Graph Representation Space for Bridging {GNN}s and {MLP}s},
author={Ling Yang and Ye Tian and Minkai Xu and Zhongyi Liu and Shenda Hong and Wei Qu and Wentao Zhang and Bin CUI and Muhan Zhang and Jure Leskovec},
booktitle={The Twelfth International Conference on Learning Representations},
year={2024}
}

@inproceedings{hu2021graph,
  title     = {GraphDIVE: Graph Classification by Mixture of Diverse Experts},
  author    = {Hu, Fenyu and Wang, Liping and Liu, Qiang and Wu, Shu and Wang, Liang and Tan, Tieniu},
  booktitle = {Proceedings of the Thirty-First International Joint Conference on
               Artificial Intelligence, {IJCAI-22}},
  pages     = {2080--2086},
  year      = {2022}
}

@article{chen2024text,
  title={Text-space graph foundation models: Comprehensive benchmarks and new insights},
  author={Chen, Zhikai and Mao, Haitao and Liu, Jingzhe and Song, Yu and Li, Bingheng and Jin, Wei and Fatemi, Bahare and Tsitsulin, Anton and Perozzi, Bryan and Liu, Hui and others},
  journal={Advances in Neural Information Processing Systems},
  volume={37},
  pages={7464--7492},
  year={2024}
}

@inproceedings{li2024sphere,
  title={Sphere: Expressive and interpretable knowledge graph embedding for set retrieval},
  author={Li, Zihao and Ao, Yuyi and He, Jingrui},
  booktitle={Proceedings of the 47th International ACM SIGIR Conference on Research and Development in Information Retrieval},
  pages={2629--2634},
  year={2024}
}

@inproceedings{shang2019end,
  title={End-to-end structure-aware convolutional networks for knowledge base completion},
  author={Shang, Chao and Tang, Yun and Huang, Jing and Bi, Jinbo and He, Xiaodong and Zhou, Bowen},
  booktitle={Proceedings of the AAAI conference on artificial intelligence},
  volume={33},
  number={01},
  pages={3060--3067},
  year={2019}
}

@article{roy2018soft1,
  title={Theory and experiments on vector quantized autoencoders},
  author={Roy, Aurko and Vaswani, Ashish and Neelakantan, Arvind and Parmar, Niki},
  journal={arXiv preprint arXiv:1805.11063},
  year={2018}
}

@inproceedings{sonderby2017soft2,
  title={Continuous relaxation training of discrete latent variable image models},
  author={S{\o}nderby, Casper Kaae and Poole, Ben and Mnih, Andriy},
  booktitle={Beysian DeepLearning workshop, NIPS},
  volume={201},
  year={2017}
}

@article{yan2024gaussian,
  title={Gaussian Mixture Vector Quantization with Aggregated Categorical Posterior},
  author={Yan, Mingyuan and Wu, Jiawei and Shah, Rushi and Liu, Dianbo},
  journal={arXiv preprint arXiv:2410.10180},
  year={2024}
}

@INPROCEEDINGS{takida2022sqvae,
    author={Takida, Yuhta and Shibuya, Takashi and Liao, WeiHsiang and Lai, Chieh-Hsin and Ohmura, Junki and Uesaka, Toshimitsu and Murata, Naoki and Takahashi, Shusuke and Kumakura, Toshiyuki and Mitsufuji, Yuki},
    title={{SQ-VAE}: Variational Bayes on Discrete Representation with Self-annealed Stochastic Quantization},
    booktitle={International Conference on Machine Learning},
    year={2022}
    }

@article{dhariwal2020jukebox,
  title={Jukebox: A generative model for music},
  author={Dhariwal, Prafulla and Jun, Heewoo and Payne, Christine and Kim, Jong Wook and Radford, Alec and Sutskever, Ilya},
  journal={arXiv preprint arXiv:2005.00341},
  year={2020}
}

@inproceedings{ramesh2021zero,
  title={Zero-shot text-to-image generation},
  author={Ramesh, Aditya and Pavlov, Mikhail and Goh, Gabriel and Gray, Scott and Voss, Chelsea and Radford, Alec and Chen, Mark and Sutskever, Ilya},
  booktitle={International conference on machine learning},
  pages={8821--8831},
  year={2021},
  organization={Pmlr}
}

@article{chang2023muse,
  title={Muse: Text-to-image generation via masked generative transformers},
  author={Chang, Huiwen and Zhang, Han and Barber, Jarred and Maschinot, AJ and Lezama, Jose and Jiang, Lu and Yang, Ming-Hsuan and Murphy, Kevin and Freeman, William T and Rubinstein, Michael and others},
  journal={arXiv preprint arXiv:2301.00704},
  year={2023}
}

@article{van2024gptvq,
  title={Gptvq: The blessing of dimensionality for llm quantization},
  author={Van Baalen, Mart and Kuzmin, Andrey and Koryakovskiy, Ivan and Nagel, Markus and Couperus, Peter and Bastoul, Cedric and Mahurin, Eric and Blankevoort, Tijmen and Whatmough, Paul},
  journal={arXiv preprint arXiv:2402.15319},
  year={2024}
}

@inproceedings{liu2025llmvq,
  title={VQ-LLM: High-performance Code Generation for Vector Quantization Augmented LLM Inference},
  author={Liu, Zihan and Luo, Xinhao and Guo, Junxian and Ni, Wentao and Zhou, Yangjie and Guan, Yue and Guo, Cong and Cui, Weihao and Feng, Yu and Guo, Minyi and others},
  booktitle={2025 IEEE International Symposium on High Performance Computer Architecture (HPCA)},
  pages={1496--1509},
  year={2025},
  organization={IEEE}
}

@inproceedings{
deng2024autoregressive,
title={Autoregressive Video Generation without Vector Quantization},
author={Haoge Deng and Ting Pan and Haiwen Diao and Zhengxiong Luo and Yufeng Cui and Huchuan Lu and Shiguang Shan and Yonggang Qi and Xinlong Wang},
booktitle={The Thirteenth International Conference on Learning Representations},
year={2025}
}

@inproceedings{
bojchevski2017deep,
title={Deep Gaussian Embedding of Graphs:  Unsupervised Inductive Learning via Ranking},
author={Aleksandar Bojchevski and Stephan Günnemann},
booktitle={International Conference on Learning Representations},
year={2018}
}

@inproceedings{mcauley2015image,
  title={Image-based recommendations on styles and substitutes},
  author={McAuley, Julian and Targett, Christopher and Shi, Qinfeng and Van Den Hengel, Anton},
  booktitle={Proceedings of the 38th international ACM SIGIR conference on research and development in information retrieval},
  pages={43--52},
  year={2015}
}

@article{shchur2018pitfalls,
  title={Pitfalls of graph neural network evaluation},
  author={Shchur, Oleksandr and Mumme, Maximilian and Bojchevski, Aleksandar and G{\"u}nnemann, Stephan},
  journal={arXiv preprint arXiv:1811.05868},
  year={2018}
}

@inproceedings{namata2012query,
  title={Query-driven active surveying for collective classification},
  author={Namata, Galileo and London, Ben and Getoor, Lise and Huang, Bert and Edu, U},
  booktitle={10th international workshop on mining and learning with graphs},
  volume={8},
  pages={1},
  year={2012}
}

@article{mialon2021graphit,
  title={Graphit: Encoding graph structure in transformers},
  author={Mialon, Gr{\'e}goire and Chen, Dexiong and Selosse, Margot and Mairal, Julien},
  journal={arXiv preprint arXiv:2106.05667},
  year={2021}
}

@inproceedings{navaneet2024compgs,
  title={Compgs: Smaller and faster gaussian splatting with vector quantization},
  author={Navaneet, KL and Pourahmadi Meibodi, Kossar and Abbasi Koohpayegani, Soroush and Pirsiavash, Hamed},
  booktitle={European Conference on Computer Vision},
  pages={330--349},
  year={2024},
  organization={Springer}
}

@inproceedings{caron2018eccv,
  title={Deep clustering for unsupervised learning of visual features},
  author={Caron, Mathilde and Bojanowski, Piotr and Joulin, Armand and Douze, Matthijs},
  booktitle={Proceedings of the European conference on computer vision (ECCV)},
  pages={132--149},
  year={2018}
}

@article{ding2021vqgnn,
  title={Vq-gnn: A universal framework to scale up graph neural networks using vector quantization},
  author={Ding, Mucong and Kong, Kezhi and Li, Jingling and Zhu, Chen and Dickerson, John and Huang, Furong and Goldstein, Tom},
  journal={Advances in Neural Information Processing Systems},
  volume={34},
  pages={6733--6746},
  year={2021}
}

@inproceedings{hou2023graphmae2,
  title={Graphmae2: A decoding-enhanced masked self-supervised graph learner},
  author={Hou, Zhenyu and He, Yufei and Cen, Yukuo and Liu, Xiao and Dong, Yuxiao and Kharlamov, Evgeny and Tang, Jie},
  booktitle={Proceedings of the ACM web conference 2023},
  pages={737--746},
  year={2023}
}

@inproceedings{
xu2018powerful,
title={How Powerful are Graph Neural Networks?},
author={Keyulu Xu and Weihua Hu and Jure Leskovec and Stefanie Jegelka},
booktitle={International Conference on Learning Representations},
year={2019}
}

@inproceedings{
velivckovic2017graph,
title={Graph Attention Networks},
author={Petar Veličković and Guillem Cucurull and Arantxa Casanova and Adriana Romero and Pietro Liò and Yoshua Bengio},
booktitle={International Conference on Learning Representations},
year={2018}
}

@article{zhang2022semi,
  title={Semi-supervised classification of graph convolutional networks with Laplacian rank constraints},
  author={Zhang, Haiqi and Lu, Guangquan and Zhan, Mengmeng and Zhang, Beixian},
  journal={Neural Processing Letters},
  volume={54},
  number={4},
  pages={2645--2656},
  year={2022},
  publisher={Springer}
}

@inproceedings{chien2021node,
  title={Node Feature Extraction by Self-Supervised Multi-scale Neighborhood Prediction},
  author={Chien, Eli and Chang, Wei-Cheng and Hsieh, Cho-Jui and Yu, Hsiang-Fu and Zhang, Jiong and Milenkovic, Olgica and Dhillon, Inderjit S},
  booktitle={International Conference on Learning Representations},
  year={2022}
}

@article{im2023vector,
  title={Vector quantization using k-means clustering neural network},
  author={Im, Sio-Kei and Chan, Ka-Hou},
  journal={Electronics Letters},
  volume={59},
  number={7},
  pages={e12758},
  year={2023},
  publisher={Wiley Online Library}
}

@inproceedings{zheng2023online,
  title={Online clustered codebook},
  author={Zheng, Chuanxia and Vedaldi, Andrea},
  booktitle={Proceedings of the IEEE/CVF International Conference on Computer Vision},
  pages={22798--22807},
  year={2023}
}

@article{dong2022denoising,
  title={Denoising aggregation of graph neural networks by using principal component analysis},
  author={Dong, Wei and Wo{\'z}niak, Marcin and Wu, Junsheng and Li, Weigang and Bai, Zongwen},
  journal={IEEE Transactions on Industrial Informatics},
  volume={19},
  number={3},
  pages={2385--2394},
  year={2022},
  publisher={IEEE}
}

@inproceedings{wu2019net,
  title={DEMO-Net: Degree-specific graph neural networks for node and graph classification},
  author={Wu, Jun and He, Jingrui and Xu, Jiejun},
  booktitle={Proceedings of the 25th ACM SIGKDD international conference on knowledge discovery \& data mining},
  pages={406--415},
  year={2019}
}

@article{van2017vqvae,
  title={Neural discrete representation learning},
  author={Van Den Oord, Aaron and Vinyals, Oriol and others},
  journal={Advances in neural information processing systems},
  volume={30},
  year={2017}
}

@article{bengio2013estimating,
  title={Estimating or propagating gradients through stochastic neurons for conditional computation},
  author={Bengio, Yoshua and L{\'e}onard, Nicholas and Courville, Aaron},
  journal={arXiv preprint arXiv:1308.3432},
  year={2013}
}

@article{wu2022nodeformer,
  title={Nodeformer: A scalable graph structure learning transformer for node classification},
  author={Wu, Qitian and Zhao, Wentao and Li, Zenan and Wipf, David P and Yan, Junchi},
  journal={Advances in Neural Information Processing Systems},
  volume={35},
  pages={27387--27401},
  year={2022}
}

@inproceedings{you2021graph,
  title={Graph contrastive learning automated},
  author={You, Yuning and Chen, Tianlong and Shen, Yang and Wang, Zhangyang},
  booktitle={International conference on machine learning},
  pages={12121--12132},
  year={2021},
  organization={PMLR}
}

@article{hu2020open,
  title={Open graph benchmark: Datasets for machine learning on graphs},
  author={Hu, Weihua and Fey, Matthias and Zitnik, Marinka and Dong, Yuxiao and Ren, Hongyu and Liu, Bowen and Catasta, Michele and Leskovec, Jure},
  journal={Advances in neural information processing systems},
  volume={33},
  pages={22118--22133},
  year={2020}
}

@article{wu2021rethinking,
  title={Rethinking infonce: How many negative samples do you need?},
  author={Wu, Chuhan and Wu, Fangzhao and Huang, Yongfeng},
  journal={arXiv preprint arXiv:2105.13003},
  year={2021}
}

@inproceedings{zeng2025hgvq,
  title={Hierarchical Vector Quantized Graph Autoencoder with Annealing-Based Code Selection},
  author={Zeng, Long and Yu, Jianxiang and Zhu, Jiapeng and Zhong, Qingsong and Li, Xiang},
  booktitle={Proceedings of the ACM on Web Conference 2025},
  pages={3772--3782},
  year={2025}
}

@article{mernyei2020wiki,
  title={Wiki-cs: A wikipedia-based benchmark for graph neural networks},
  author={Mernyei, P{\'e}ter and Cangea, C{\u{a}}t{\u{a}}lina},
  journal={arXiv preprint arXiv:2007.02901},
  year={2020}
}

@inproceedings{
platonov2023critical,
title={A critical look at the evaluation of {GNN}s under heterophily: Are we really making progress?},
author={Oleg Platonov and Denis Kuznedelev and Michael Diskin and Artem Babenko and Liudmila Prokhorenkova},
booktitle={The Eleventh International Conference on Learning Representations },
year={2023}
}

@article{zeghidour2021reset,
  title={Soundstream: An end-to-end neural audio codec},
  author={Zeghidour, Neil and Luebs, Alejandro and Omran, Ahmed and Skoglund, Jan and Tagliasacchi, Marco},
  journal={IEEE/ACM Transactions on Audio, Speech, and Language Processing},
  volume={30},
  pages={495--507},
  year={2021},
  publisher={IEEE}
}

@inproceedings{lancucki2020ema,
  title={Robust training of vector quantized bottleneck models},
  author={{\L}a{\'n}cucki, Adrian and Chorowski, Jan and Sanchez, Guillaume and Marxer, Ricard and Chen, Nanxin and Dolfing, Hans JGA and Khurana, Sameer and Alum{\"a}e, Tanel and Laurent, Antoine},
  booktitle={2020 International Joint Conference on Neural Networks (IJCNN)},
  pages={1--7},
  year={2020},
  organization={IEEE}
}

@inproceedings{huh2023straightening,
  title={Straightening out the straight-through estimator: Overcoming optimization challenges in vector quantized networks},
  author={Huh, Minyoung and Cheung, Brian and Agrawal, Pulkit and Isola, Phillip},
  booktitle={International Conference on Machine Learning},
  pages={14096--14113},
  year={2023},
  organization={PMLR}
}

@inproceedings{
yu2021ortho,
title={Vector-quantized Image Modeling with Improved {VQGAN}},
author={Jiahui Yu and Xin Li and Jing Yu Koh and Han Zhang and Ruoming Pang and James Qin and Alexander Ku and Yuanzhong Xu and Jason Baldridge and Yonghui Wu},
booktitle={International Conference on Learning Representations},
year={2022}
}

@inproceedings{
velivckovic2018dgi,
title={Deep Graph Infomax},
author={Petar Veličković and William Fedus and William L. Hamilton and Pietro Liò and Yoshua Bengio and R Devon Hjelm},
booktitle={International Conference on Learning Representations},
year={2019}
}

@inproceedings{chen2021incremental,
  title={Incremental few-shot learning via vector quantization in deep embedded space},
  author={Chen, Kuilin and Lee, Chi-Guhn},
  booktitle={International conference on learning representations},
  year={2021}
}

@article{polyak1992acceleration,
  title={Acceleration of stochastic approximation by averaging},
  author={Polyak, Boris T and Juditsky, Anatoli B},
  journal={SIAM journal on control and optimization},
  volume={30},
  number={4},
  pages={838--855},
  year={1992},
  publisher={SIAM}
}

@article{lu2023hierarchical,
  title={Hierarchical vector quantized transformer for multi-class unsupervised anomaly detection},
  author={Lu, Ruiying and Wu, YuJie and Tian, Long and Wang, Dongsheng and Chen, Bo and Liu, Xiyang and Hu, Ruimin},
  journal={Advances in Neural Information Processing Systems},
  volume={36},
  pages={8487--8500},
  year={2023}
}

@inproceedings{tang2022avqvc,
  title={Avqvc: One-shot voice conversion by vector quantization with applying contrastive learning},
  author={Tang, Huaizhen and Zhang, Xulong and Wang, Jianzong and Cheng, Ning and Xiao, Jing},
  booktitle={ICASSP 2022-2022 IEEE International Conference on Acoustics, Speech and Signal Processing (ICASSP)},
  pages={4613--4617},
  year={2022},
  organization={IEEE}
}

@inproceedings{
fifty2024restructuring,
title={Restructuring Vector Quantization with the Rotation Trick},
author={Christopher Fifty and Ronald Guenther Junkins and Dennis Duan and Aniketh Iyengar and Jerry Weihong Liu and Ehsan Amid and Sebastian Thrun and Christopher Re},
booktitle={The Thirteenth International Conference on Learning Representations},
year={2025}
}

@inproceedings{
thakoor2021bgrl,
title={Large-Scale Representation Learning on Graphs via Bootstrapping},
author={Shantanu Thakoor and Corentin Tallec and Mohammad Gheshlaghi Azar and Mehdi Azabou and Eva L Dyer and Remi Munos and Petar Veli{\v{c}}kovi{\'c} and Michal Valko},
booktitle={International Conference on Learning Representations},
year={2022}
}

@inproceedings{hou2022graphmae,
  title={Graphmae: Self-supervised masked graph autoencoders},
  author={Hou, Zhenyu and Liu, Xiao and Cen, Yukuo and Dong, Yuxiao and Yang, Hongxia and Wang, Chunjie and Tang, Jie},
  booktitle={Proceedings of the 28th ACM SIGKDD conference on knowledge discovery and data mining},
  pages={594--604},
  year={2022}
}

@article{rampavsek2022graphgps,
  title={Recipe for a general, powerful, scalable graph transformer},
  author={Ramp{\'a}{\v{s}}ek, Ladislav and Galkin, Michael and Dwivedi, Vijay Prakash and Luu, Anh Tuan and Wolf, Guy and Beaini, Dominique},
  journal={Advances in Neural Information Processing Systems},
  volume={35},
  pages={14501--14515},
  year={2022}
}

@article{wu2023sgformer,
  title={Sgformer: Simplifying and empowering transformers for large-graph representations},
  author={Wu, Qitian and Zhao, Wentao and Yang, Chenxiao and Zhang, Hengrui and Nie, Fan and Jiang, Haitian and Bian, Yatao and Yan, Junchi},
  journal={Advances in Neural Information Processing Systems},
  volume={36},
  pages={64753--64773},
  year={2023}
}

@inproceedings{shirzad2023exphormer,
  title={Exphormer: Sparse transformers for graphs},
  author={Shirzad, Hamed and Velingker, Ameya and Venkatachalam, Balaji and Sutherland, Danica J and Sinop, Ali Kemal},
  booktitle={International Conference on Machine Learning},
  pages={31613--31632},
  year={2023},
  organization={PMLR}
}

@article{chung2020vector,
  title={Vector-quantized autoregressive predictive coding},
  author={Chung, Yu-An and Tang, Hao and Glass, James},
  journal={arXiv preprint arXiv:2005.08392},
  year={2020}
}

@inproceedings{zhang1997improvement,
  title={An improvement image vector quantization based on affine transformation},
  author={Zhang, Ying and Yu, Ying-Lin and Po, Lai-Man},
  booktitle={1997 IEEE International Conference on Systems, Man, and Cybernetics. Computational Cybernetics and Simulation},
  volume={2},
  pages={1094--1099},
  year={1997},
  organization={IEEE}
}

@article{wu2019vector,
  title={Vector quantization: a review},
  author={Wu, Ze-bin and Yu, Jun-qing},
  journal={Frontiers of Information Technology \& Electronic Engineering},
  volume={20},
  number={4},
  pages={507--524},
  year={2019},
  publisher={Springer}
}

@article{williams2020hierarchical,
  title={Hierarchical quantized autoencoders},
  author={Williams, Will and Ringer, Sam and Ash, Tom and MacLeod, David and Dougherty, Jamie and Hughes, John},
  journal={Advances in Neural Information Processing Systems},
  volume={33},
  pages={4524--4535},
  year={2020}
}
\bibliographystyle{icml2026}

\newpage
\appendix
\onecolumn

\begin{table*}[t]
\caption{Dataset statistics for selected datasets.}
\resizebox{\textwidth}{!}{
\centering
\begin{tabular}{l l l r r r r r}
\toprule
\textbf{Dataset} & \textbf{Domain} & \textbf{Task} & \textbf{\# Graphs} & \textbf{Avg. \#Nodes} & \textbf{Avg. \#Edges} & \textbf{\# Classes} & { \textbf{Metric}}\\
\midrule
CiteSeer        & Citation     & Node  & 1       & 3,327     & 4,522      & 6    & { Accuracy} \\
Cora            & Citation     & Node  & 1       & 2,708     & 10,556     & 7    & { Accuracy} \\
PubMed          & Citation     & Node  & 1       & 19,717    & 88,651     & 3   & { Accuracy}  \\
Computer        & Co-purchase  & Node  & 1       & 13,752    & 491,722    & 10   & { Accuracy} \\
Photo           & Co-purchase  & Node  & 1       & 7,650     & 238,163    & 8   & { Accuracy}  \\
WikiCS          & Web link     & Node  & 1       & 11,701    & 216,123    & 10  & { Accuracy}  \\
Amazon-Ratings  & Review       & Node  & 1       & 22,662    & 32,927     & 18   & { Accuracy} \\
Roman-Empire    & Synthetic     & Node  & 1       & 24,492    & 93,050     & 5   & { Accuracy}  \\
Questions       & Synthetic     & Node  & 1       & 48,921    & 153,540    & 2   & { ROC-AUC}  \\
FB15K237        & Knowledge    & Link  & 1       & 14,541    & 310,116    & 237 & { Accuracy}  \\
WN18RR          & Knowledge    & Link  & 1       & 40,943    & 93,003     & 11  & { Accuracy}  \\
PCBA            & Molecule     & Graph & 437,929 & 26.0      & 28.1       & 128  & { ROC-AUC} \\
HIV             & Molecule     & Graph & 41,127  & 25.5      & 27.5       & 2   & { ROC-AUC}  \\
\bottomrule
\end{tabular}
}
\label{tab:merged_dataset_statistics}
\end{table*}
\section{Experimental Setup}\label{appendix:experiments}
\subsection{Dataset}
We use both homophilous and heterophilous graphs in our experiments.
To implement empirical study and evaluate codebook utilization, we use various datasets, including Cora~\cite{bojchevski2017deep}, CiteSeer, PubMed~\cite{namata2012query}, Amazon-Computer, Amazon-Photo~\cite{shchur2018pitfalls,mcauley2015image}, WikiCS~\cite{mialon2021graphit}, Amazon-Ratings~\cite{platonov2023critical}, and Roman-Empire~\cite{platonov2023critical}.
To assess transferability, we use cross-task and cross-domain datasets. Specifically, we use Cora, PubMed, and WikiCS for node classification; WN18RR~\cite{shang2019end} and FB15K237~\cite{li2024sphere} for link prediction; and HIV~\cite{hu2021graph} and PCBA~\cite{chen2024text} for graph classification.
Finally, we evaluate serialization ability using the same datasets employed for codebook utilization.
Detailed dataset statistics are summarized in Table~\ref{tab:merged_dataset_statistics}.
\subsection{Baseline}
We use different baselines for the empirical study and three parts of our main experiments.

\myparagraph{Empirical Study and Codebook Utilization}
We primarily adopt codebook mitigation strategies originally developed in the vision and language domains, including EMA~\cite{lancucki2020ema}, affine parameters~\cite{huh2023straightening}, codebook reset~\cite{zeghidour2021reset}, and pretrained encoders~\cite{zhao2024pretrainVQ}.
    We further include SimVQ~\cite{zhu2024softassign}, which addresses the codebook collapse via one-single MLP layer over the latent basis vectors.
    Additionally, we compare with HQA-GAE~\cite{zeng2025hgvq}, a recent graph VQ model that applies a hierarchical VQ structure and an annealing strategy for codeword selection.
    
\myparagraph{Transferability} 
To evaluate the effectiveness of RGVQ in learning transferrable graph tokens, we integrate it into a graph foundation model, i.e., GFT, and compare it with vanilla Graph VQ and its variants with different mitigation strategies. 
    Moreover, we include supervised GNNs, i.e., GCN, GAT, and GIN, and graph self-supervised methods, i.e., DGI~\cite{velivckovic2018dgi}, BGRL~\cite{thakoor2021bgrl}, GraphMAE~\cite{hou2022graphmae}, and GAINT~\cite{chien2021node}. 
    The supervised GNNs are trained directly on each target dataset, while the self-supervised methods and all GFT variants are pretrained on the full set of datasets and then fine-tuned per target.
    
\myparagraph{Serialization}
To further evaluate the effectiveness of RGVQ in serialization, we integrate it into the Graph Quantized Transformer (GQT)~\cite{wang2025learning}, where discrete tokens serve as the input sequence to a vanilla Transformer backbone. 
    We follow the original sequence reconstruction method and compare the performance of RGVQ-enhanced GQT against the original GQT, supervised GNNs, and graph transformers, including GraphGPS~\cite{rampavsek2022graphgps}, SGFormer~\cite{wu2023sgformer}, Exphomer~\cite{shirzad2023exphormer}, and Nodeformer~\cite{wu2022nodeformer}.

\subsection{Implementation Details}
\myparagraph{Empirical Study}
We provide the hyperparameters and experimental setup used in the empirical study of codebook perplexity.
We jointly train the single-head VQ model and the GAT encoder using the link prediction and feature reconstruction tasks, along with the commitment loss and vocabulary loss. 
The task weights are set to 0.01, 100, 0.1, and 0.9, respectively.
We train the model for 1000 epochs and report the highest perplexity during the training process for each method and a specific codebook size $K$.
For all methods, we utilize the kmeans initialization and orthogonal regulation for the codebook, with a regularization weight of 0.1.
The GNN consists of 4 layers with a hidden dimension of 256.
AdamW is utilized as the optimizer with a learning rate of 1e-4 and a weight decay of 1e-5.
For affine parameters, we use Euclidean distance and set the codebook decay to 0.9.
For codebook reset, the threshold of deadcode is set to 10.
For pretraining encoder, we pretrain the GNN encoder for 50 epochs before the joint training.

\myparagraph{Codebook Utilization} 
    The implementation of all collapse mitigation strategies is the same with the empirical study.
    For HQA-GAE, we use one-head VQ model and use the same hidden dimension and number of GNN layers as other methods, while all remaining hyperparameters follow the original paper.
    For RGVQ, we set the codebook size $K$ to 512. 
    To construct the contrastive sample sets, for each node, we construct a pool of positive candidates by combining its 1-hop neighbors with the nodes that are most 20 similar in the input feature space (top-$K=20$), which is measured by cosine similarity. From this pool, we sample 50 nodes with replacement as positive samples. The negative pool is defined symmetrically as all nodes that are neither neighbors nor feature positives, and we sample 50 negative nodes with replacement from the negative pool.
    The training weights for the link reconstruction, node feature reconstruction, contrastive regularization, commitment loss, and vocabulary loss are set to 0.01, 100, 1, 0.1, and 0.9, respectively.
    We set the temperature for the Gumbel-Softmax trick to 0.1. 
    { {We train on each dataset for 1000 epochs to ensure convergence, and repeat the process 20 times to report the mean perplexity with standard deviations.}}
    Detailed hyperparameters for each dataset are summarized in Table~\ref{tab:hyperparams}.
\begin{table*}[t]
\centering
\setlength{\tabcolsep}{5pt}
\renewcommand{\arraystretch}{1.2}
\caption{{ Hyperparameters of RGVQ for each dataset.}}
\resizebox{0.98\textwidth}{!}{
\begin{tabular}{lccccccccc}
\hline
\textbf{Hyperparameter} 
& \textbf{Cora} & \textbf{Pubmed} & \textbf{Citeseer}
& \textbf{Computer} & \textbf{Photo} & \textbf{WikiCS}
& \textbf{Ratings} & \textbf{Roman} & \textbf{Questions} \\
\hline
Hidden dimension                & 256 & 256 & 256 & 256 & 256 & 256 & 256 & 256 & 256 \\
Learning rate             & 0.001 & 0.001 & 0.001 & 0.001 & 0.001 & 0.001 & 0.001 & 0.001 & 0.001 \\
Weight decay              & 1e-5 & 1e-5 & 1e-5 & 1e-5 & 1e-5 & 1e-5 & 1e-5 & 1e-5 & 1e-5 \\
Seed                      & 42  & 42  & 42  & 42  & 42  & 42  & 42  & 42  & 42 \\
Epochs                    & 1000 & 1000 & 1000 & 1000 & 1000 & 1000 & 1000 & 1000 & 1000 \\
Feature Reconstruction    & 100 & 100 & 100 & 100 & 100 & 100 & 100 & 100 & 100 \\
Topology Reconstruction   & 0.01 & 0.01 & 0.01 & 0.01 & 0.01 & 0.01 & 0.01 & 0.01 & 0.01 \\
$\beta$                   & 1 & 1 & 1 & 1 & 1 & 1 & 1 & 1 & 1 \\
Temperature               & 0.1 & 0.1 & 0.1 & 0.1 & 0.1 & 0.1 & 0.1 & 0.1 & 0.1 \\
Similarity function       & Cosine & Cosine & Cosine & Cosine & Cosine & Cosine & Cosine & Cosine & Cosine \\
Top-$K$                   & 20 & 20 & 20 & 20 & 20 & 20 & 20 & 20 & 20 \\
Sample number             & 50 & 50 & 50 & 50 & 50 & 50 & 50 & 50 & 50 \\
GNN layers                & 4 & 4 & 4 & 4 & 4 & 4 & 4 & 4 & 4 \\
\hline
\end{tabular}
}
\label{tab:hyperparams}
\end{table*}
\begin{table*}[t]
\centering
\caption{{ Selected hyperparameters in GQT for each dataset.}}
\resizebox{0.98\textwidth}{!}{
\begin{tabular}{lcccccccccc}
\toprule
& \multicolumn{2}{c}{GNN Encoder} & \multicolumn{2}{c}{Quantizer} & \multicolumn{5}{c}{Transformer} \\
\cmidrule(lr){2-3} \cmidrule(lr){4-5} \cmidrule(lr){6-10}
Dataset & \# layers & \# Hidden dim & \# Codebooks & Codebook size & KNN & PPR & \# Layers & \# Heads & \# FFN dim \\
\midrule
Cora        & 2 & 256 & 3 & 128 & 0 & 15 & 2 & 4 & 512 \\
CiteSeer        & 2 & 256 & 3 & 128 & 5 & 15 & 2 & 4 & 512 \\
PubMed          & 2 & 256 & 3 & 256 & 0 & 15 & 2 & 4 & 512 \\
Computer        & 2 & 256 & 3 & 128 & 5 & 30 & 2 & 4 & 512 \\
Photo           & 3 & 512 & 3 & 128 & 5 & 20 & 2 & 4 & 1024 \\
WikiCS          & 2 & 256 & 3 & 128 & 5 & 30 & 2 & 4 & 512 \\
Amazon-Ratings  & 4 & 512 & 3 & 128 & 5 & 20 & 2 & 4 & 1024 \\
Roman-Empire    & 6 & 256 & 3 & 256 & 10 & 15 & 3 & 4 & 512 \\
Questions       & 3 & 256 & 3 & 512 & 10 & 15 & 2 & 4 & 512 \\
\bottomrule
\end{tabular}
}
\label{table:GQT para}
\end{table*}

\myparagraph{Transferability} We use RGVQ as a plugin within the pretraining pipeline of GFT.  
{ Specifically, we retain the same pretraining tasks in GFT~\cite{wang2024gft}, including the link, node feature, and node embedding reconstruction tasks, and integrate RGVQ as a regularization term.
Their weights are set to 100, 1, 0.01, and 10, respectively.
For the backbone encoder, we utilize a 2-layer GCN model with ReLU activation, and set the codebook size to 512 and the hidden dimension to 256.
We use AdamW optimizer with a learning rate of 1e-3 and weight decay of 1e-5. 
For data augmentation, we apply a link drop rate and the node-feature drop rate of 0.2. 
We pretrain the VQ tokens for 500 epochs on all datasets.
During finetuning, we repeat each experiment 20 times to report the average performance with standard deviations.
We finetune the model for 250 epochs using early stopping.
For dataset splits, we follow the commonly used protocol for Cora and PubMed and utilize the predefined 10 splits with different seeds to report the downstream performance. 
Each split includes 20 labeled nodes per class for training.
For WikiCS, we follow the recommended protocol by OGB and use the official split, reporting average performance across 20 splits~\cite{mernyei2020wiki}.
For WN18RR, we utilize 86,835/3,034/3,134 links for training/validation/test, respectively.
For FB15K237, we use 272,155/17,535/20,466 links for training/validation/test, respectively.
For HIV and PCBA, we follow the official data split and utilize 80\%/10\%/10\% for training/validation/test set~\cite{hu2020open}.}
    
\myparagraph{Serialization}
{ The training of GQT includes two parts: the VQ tokenizer and the backbone transformer. We detail the implementations and training hyperparameters below.
For the VQ tokenizer, we follow the original paper and use Residual VQ~\cite{wang2025learning}. 
We retain all of the reconstruction tasks in the pretraining setting of GQT, including Deep Graph Infomax (DGI)~\cite{velivckovic2018dgi} and GraphMAE2~\cite{hou2023graphmae2}, and integrate RGVQ as a regularization term.
For the tokenizer, we set the number of codebooks to three for GQT, GQT + EMA, GQT + AP, GQT + Reset, GQT + PT; and one for GQT + SimVQ, GQT + RGVQ.
We choose codebook size from \{128,256,512\}.
For the GNN encoder, we adopt GCN with ReLU activation, varying the number of layers from \{2,3,4,6\} and hidden dimensions from \{256,512\}.
We pretrain the VQ tokenizer and the GNN encoder for 200 epochs until convergence.
For training the vanilla transformer, we construct semantic links using K-Nearest-Neighbors, with K in \{0,5,10\}.
To serialize the input graph sequence, we use Personalized PageRank (PPR) to generate a sequence for each node, with the sequence length selected from \{15,20,30\}.
The transformer uses 2 or 3 layers, 4 attention heads, and a feedforward dimension of \{512, 1024\}.
The detailed hyperparameters are summarized in Table~\ref{table:GQT para}.
We train transformers with node labels together with the pretrained VQ tokenizer and GNN encoder, and report the average performance and standard deviations over 5 runs.
For Cora, Pubmed, Citeseer, Computer, Photo, we follow the original settings in~\cite{wang2025learning}, using 60\%/20\%/20\% for training/validation/test.
For WikiCS, we follow the predefined split in~\cite{mernyei2020wiki} and report the average performance across 20 splits.
For Amazon-Ratings, Roman-Empire, and Questions, we adopt the splits in~\cite{platonov2023critical}, using 50\%/25\%/25\% for training/validation/test, and report the mean performance over 10 random splits. } 
\begin{figure*}[t]
  \centering
\includegraphics[width=1.0\linewidth]{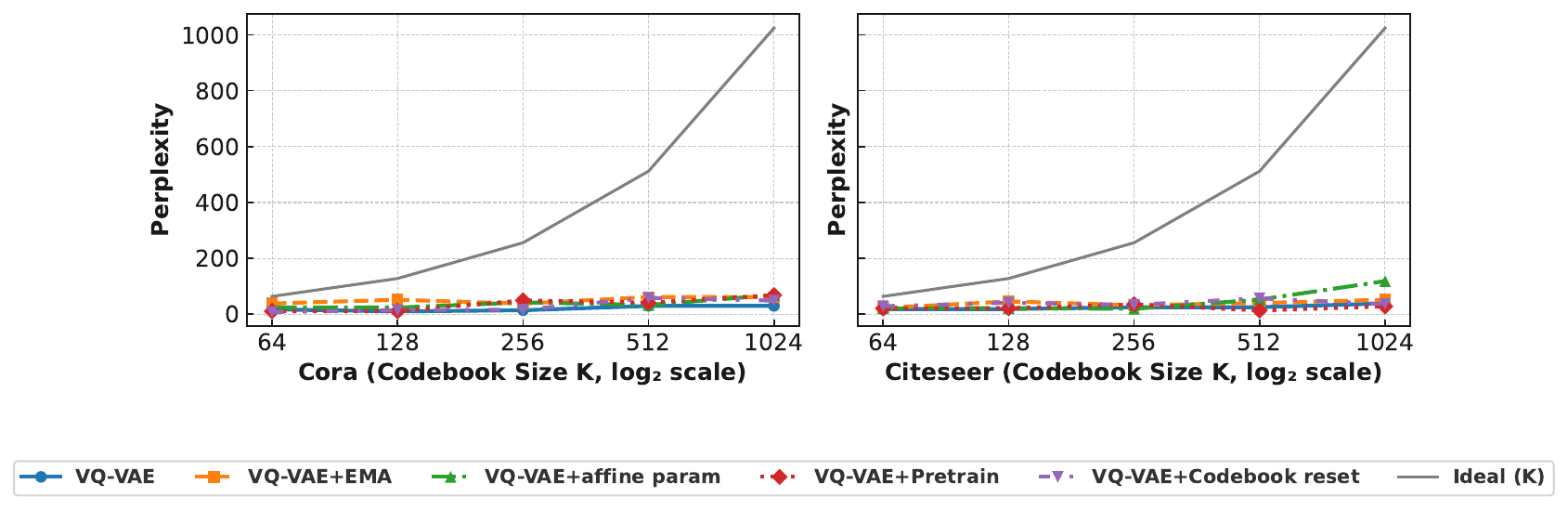}
\caption{Codebook perplexites on graph datasets. The black lines indicate the optimal perplexities, i.e., codebook size K.}
  \label{fig:supp_perplexity}
\end{figure*}

\begin{figure*}[t]
  \centering
  \begin{subfigure}[t]{0.32\linewidth}
    \centering
    \includegraphics[width=0.95\linewidth]{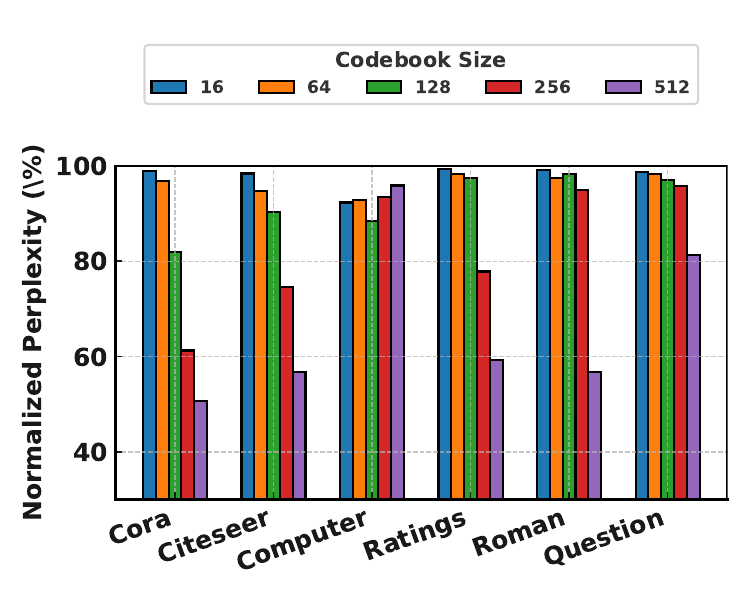}
    \caption{Codebook size vs. Perplexity.}
    \label{fig:codebook size}
  \end{subfigure}
  \hfill
  \begin{subfigure}[t]{0.32\linewidth}
    \centering
    \includegraphics[width=0.95\linewidth]{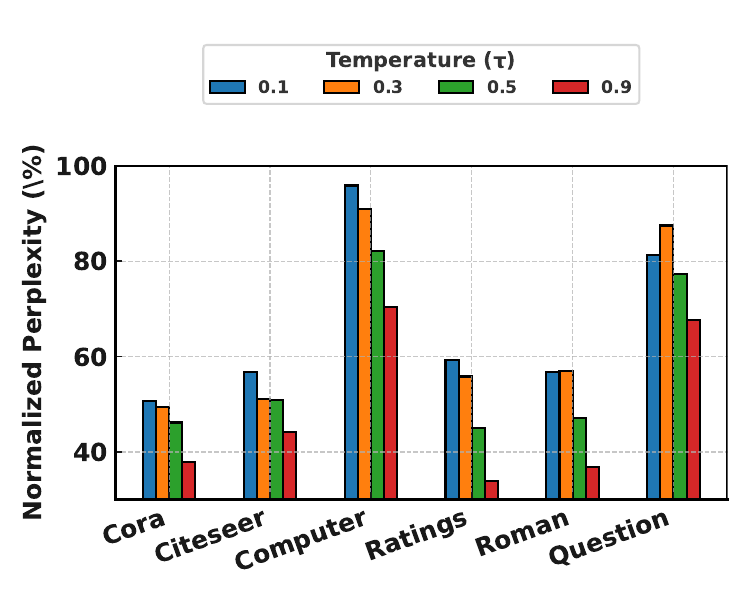}
    \caption{Temperature $(\tau)$ vs. Perplexity.}
    \label{fig:add_temp}
  \end{subfigure}
  \hfill
  \begin{subfigure}[t]{0.32\linewidth}
    \centering
    \includegraphics[width=0.95\linewidth]{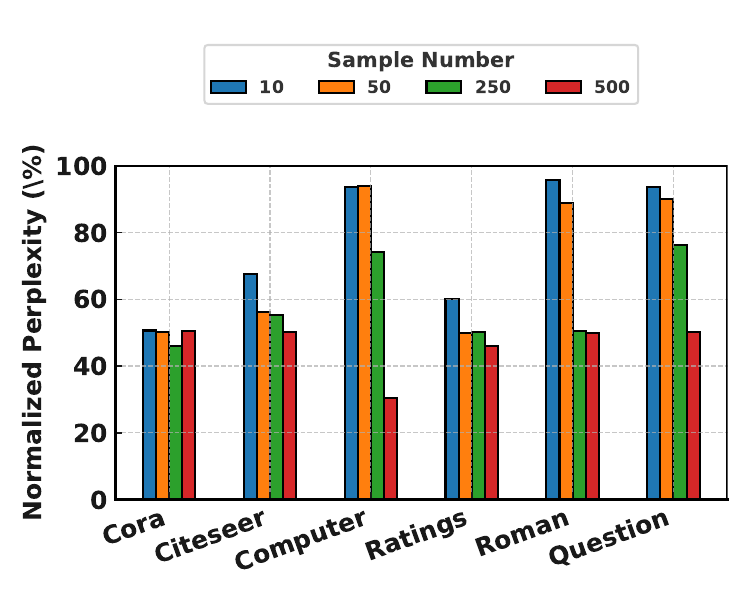}
    \caption{Sample Number vs. Perplexity.}
    \label{fig:ladd_sample}
  \end{subfigure}
  \caption{Ablation study results.}
  \label{fig:ablation}
\end{figure*}
\begin{figure*}[t]
    \centering
    \includegraphics[width=0.9\linewidth]{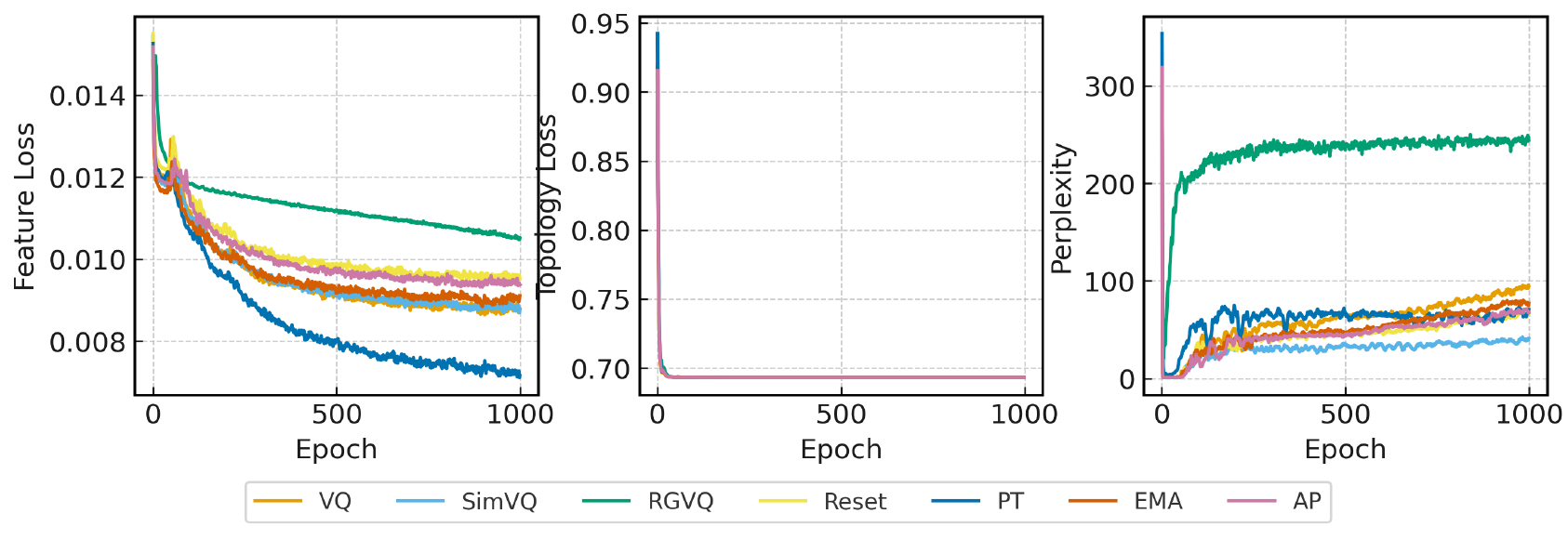}
    \caption{{ Reconstruction loss and perplexity during the pretraining process on Cora}}
    \label{pretraining process}
\end{figure*}    
\section{Additional Experiments}\label{appendix:additional}
\myparagraph{Empirical Study}
We additionally provide the quantization results on Cora and Citeseer datasets.
The results shown in Figure~\ref{fig:supp_perplexity} suggest that codebook collapse is a systematic problem in Cora and Citeseer datasets, even though the mitigation strategies are applied.

\begin{table}[t]
\centering
\caption{Dataset statistics (PCA@95, average degree) and measured codebook perplexity across 8 graph datasets.}
\begin{tabular}{lccc}
\toprule
\textbf{Dataset} & \textbf{PCA@95} & \textbf{Avg Degree} & \textbf{Perplexity} \\
\midrule
Cora       & 802  & 4.90  & 94.47 \\
Pubmed    & 410  & 5.50  & 4.14  \\
Citeseer  & 1459 & 3.74  & 60.09 \\
Photo     & 611  & 32.13 & 1.00  \\
Computer  & 646  & 36.76 & 1.00  \\
Ratings   & 194  & 8.60  & 13.29 \\
Roman     & 141  & 3.91  & 10.84 \\
Questions & 160  & 7.28  & 20.78 \\
\bottomrule
\end{tabular}
\label{tab:instance}
\end{table}

\begin{table}[t]
\centering
\caption{Multivariate regression of PCA@95, average degree, and perplexity.}
\begin{tabular}{lcc}
\toprule
 & \textbf{PCA@95} & \textbf{Avg Degree} \\
\midrule
Pearson $r$ (p-value) & 0.714 (0.013) & -0.249 (0.461) \\
Partial $r$ (p-value) & 0.733 (0.010) & -0.336 (0.312) \\
$\beta$ (p-value) & 0.026 (0.016) & -0.230 (0.342) \\
\bottomrule
\end{tabular}
\label{tab:correlation_results}
\end{table}
\myparagraph{{Codebook diversity on real datasets}}
We use PCA@95 as a proxy for effective feature redundancy and the average node degree as a proxy for local connectivity. 
We provide statistics on eight real graphs in Table~\ref{tab:instance}, from which we observe that datasets with higher average degree (Photo, Computer) or lower PCA@95 (Ratings, Roman, Questions) exhibit lower perplexity, whereas datasets with both higher PCA@95 and lower degree (Cora, Citeseer) exhibit much weaker collapse. Additionally, we perform a multivariate regression where perplexity is modeled as a function of PCA@95 and Avg. degree. We report regression coefficients ($\beta$), Pearson coefficients, partial correlations between each variable and perplexity, and their p-values in Table~\ref{tab:correlation_results}. PCA@95 shows a positive correlation, while Avg. degree exhibits a negative trend but does not reach statistical significance. These results are consistent with the trends observed in Figure~\ref{fig:property_vs_perplexity1}. Because the analysis is based on limited datasets, which may constrain the statistical power, we treat this as empirical evidence rather than a causal conclusion.

\begin{table*}[t]
\centering

\caption{{ Perplexity with varying number of GNN layers $L$.}}
\resizebox{0.98\textwidth}{!}{
\begin{tabular}{l|cc|cc|cc|cc|cc}
\toprule
\multirow{2}{*}{Dataset} &
\multicolumn{2}{c|}{$L=1$} &
\multicolumn{2}{c|}{$L=2$} &
\multicolumn{2}{c|}{$L=3$} &
\multicolumn{2}{c|}{$L=4$} &
\multicolumn{2}{c}{$L=5$} \\
\cmidrule(lr){2-11}
 & VQ & RGVQ & VQ & RGVQ & VQ & RGVQ & VQ & RGVQ & VQ & RGVQ \\
\midrule
Cora      & 154.34 & 394.96 & 121.59 & 339.19 & 109.06 & 257.47 & 94.47 & 211.69 & 99.44 & 218.45 \\
Pubmed    & 8.97   & 452.16 & 3.12   & 300.51 & 5.18   & 295.64 & 4.14  & 319.09 & 4.07  & 295.64 \\
Photo     & 1.99   & 432.32 & 1.00   & 421.04 & 1.00   & 443.60 & 1.00  & 446.02 & 1.00  & 306.06 \\
Computer  & 3.81   & 468.98 & 1.00   & 452.41 & 1.00   & 464.65 & 1.00  & 413.10 & 1.00  & 394.83 \\
Ratings   & 32.64  & 414.10 & 15.59  & 295.28 & 10.80  & 213.42 & 13.29 & 200.93 & 9.14  & 207.18 \\
\bottomrule
\end{tabular}}
\label{tab: gnn layer}
\end{table*}

\myparagraph{Ablation Study}
We also provide the additional ablation study to further evaluate the contribution of each proposed module in RGVQ in this section.
Here we use the normalized perplexity, defined as the ratio of utilized codebook entries to the total size of the codebook.
First, we evaluate how codebook size affects the codebook utilization of RGVQ. 
The results are shown in Figure~\ref{fig:codebook size}. 
Across all datasets, RGVQ consistently maintains high normalized perplexity as the codebook size increases, showing strong robustness to the choice of codebook size.  
Notably, even with a large codebook size of 512, the model utilizes over 50\% of the codebook capacity across all datasets.  
Second, we investigate how the Gumbel–Softmax temperature $\tau$ affects normalized perplexity. 
A lower temperature yields a codebook assignment distribution that is closer to one-hot.
As shown in Figure~\ref{fig:add_temp}, unlike some prior work~\cite{zeng2025hgvq} that rely on temperature annealing, a relatively low and fixed temperature is sufficient to address the non-differentiability of deterministic VQ on selected datasets.
Finally, we examine how the number of contrastive samples in structure-aware regularization affects normalized perplexity.
As shown in Figure~\ref{fig:ladd_sample}, even a small number of contrastive samples (e.g., $n=10$ or $50$) achieves relatively high normalized perplexity, indicating effective codebook use.
Increasing $n$ does not consistently lead to better utilization and may even degrade in some datasets, potentially due to added training noise.

\myparagraph{Converge Analysis}
We also provide the reconstruction loss and perplexity curves during the pretraining process of RGVQ and all baselines with codebook size $K=512$.
As shown in Figure~\ref{pretraining process}, all baselines reach stable reconstruction loss within the first 250 epochs and remain stable afterwards, while they all collapse to less than 100 and do not recover. 
This confirms that the collapse is a problem for vanilla VQ and other anti-collapse solutions.
Regarding reconstruction performance, collapse does not necessarily produce large reconstruction losses because a strong decoder can overfit to node features or links even though usable tokens are limited. 
However, this phenomenon is fundamentally undesirable. 
When nodes collapse to the small portion of tokens, the discrete latent space becomes degenerate and ceases to reflect any structural or semantic diversity in the graph. In this situation, the VQ module fails to provide meaningful discrete representations. 
While the appropriate codebook size may depend on task-specific trade-offs between compression and expressiveness, our method offers a flexible and effective framework that preserves token diversity while scaling to larger codebook sizes.

\myparagraph{Influence of GNN Layer Number}
To better understand the relations between GNN layer number and quantization diversity, we evaluate how the number of GNN layer $L$ affects the quantization perplexity in RGVQ and vanilla VQ. Based on the results in Table~\ref{tab: gnn layer}, we make the following observations: (1) As the layer number $L$ increases, the perplexity of vanilla VQ consistently decreases. Deeper GNNs suffer from over-smoothing, causing node representations to fall more easily within the radius of the same codeword, resulting in less diverse quantization. 
(2) RGVQ is robust for different layers because it provides explicit regularization. 

\section{Complexity Analysis}\label{appendix:complexity}
Assume a $L$-layer GNN, a codebook of size $K$, and hidden dimension of $d$, the number of nodes and links are denoted as $|\mathcal{V}|$ and $|\mathcal{E}|$ respectively.
We divide the complexity analysis into two parts: Pre-computation of contrastive set and quantization process.

\myparagraph{Pre-computation of Contrastive Set} 
Before training, RGVQ constructs for each node sets of positive and negative samples, based on both structural and feature similarity. This step is performed once and reused during training.
To implement this, neighbors are first extracted by scanning the adjacency matrix, which requires $O(|\mathcal{E}|)$ time. 
Then compute feature distances between each node and all others will take $O(|\mathcal{V}|^2 d)$ time. 
However, in practice, we adopt a sampling strategy: for each node, we sample $M$ non-neighbor nodes (where $M$ is a small constant, e.g., 100) and compute their feature similarity. 
This limits the total cost of semantic similarity computation to $O(|\mathcal{V}| \cdot M \cdot d)$, which is linear in the number of nodes.
After collecting both structurally and semantically similar candidates, we perform top-$k$ selection for each node to finalize its positive sample set, costing $O(|\mathcal{V}| \log k)$ time in total. 
Negative pairs are sampled from the set of all nodes excluding the positives.
Thus, the overall time complexity of the contrastive set construction process is 
$O(|\mathcal{E}| + |\mathcal{V}| \cdot M \cdot d + |\mathcal{V}| \log k)$, which is linear in the number of nodes and edges under fixed $M$ and $k$. 
\begin{table*}[t]
\centering
\caption{Performance on large graphs.}
\resizebox{0.98\textwidth}{!}{
\begin{tabular}{lccccccc}
\toprule
Dataset & VQ & EMA & AP & Reset & PT & SimVQ & RGVQ \\
\midrule
Arxiv 
& \makecell{6.43 $\pm$ 0.94}
& \makecell{7.42 $\pm$ 1.13}
& \makecell{52.39 $\pm$ 5.56}
& \makecell{121.24 $\pm$ 9.46}
& \makecell{7.85 $\pm$ 1.05}
& \makecell{54.34 $\pm$ 7.03}
& \makecell{\textbf{305.13} $\pm$ 20.30} \\

Products 
& \makecell{10.13 $\pm$ 1.05}
& \makecell{13.45 $\pm$ 1.79}
& \makecell{67.71 $\pm$ 7.84}
& \makecell{105.23 $\pm$ 15.45}
& \makecell{8.79 $\pm$ 0.98}
& \makecell{61.12 $\pm$ 13.23}
& \makecell{\textbf{223.47} $\pm$ 10.81} \\

Proteins 
& \makecell{9.85 $\pm$ 0.86}
& \makecell{14.87 $\pm$ 1.42}
& \makecell{49.32 $\pm$ 7.51}
& \makecell{123.74 $\pm$ 20.01}
& \makecell{8.47 $\pm$ 0.82}
& \makecell{56.63 $\pm$ 9.72}
& \makecell{\textbf{254.87} $\pm$ 10.67} \\
\bottomrule
\end{tabular}}
\label{tab:ogb_results}
\end{table*}

\myparagraph{Quantization Process} 
The time and space complexity of the GNN encoder are $O(L d^2 |\mathcal{V}| + L d |\mathcal{E}|)$ and $O(L d^2 + L d |\mathcal{V}| + |\mathcal{E}|)$, respectively. The decoder has the same complexity.
RGVQ computes the distance between each node embedding and all $K$ codewords, and estimates the soft assignment distribution via the Gumbel-Softmax trick. This process requires $O(|\mathcal{V}| K d)$ time and $O(|\mathcal{V}| K)$ space.
Finally, RGVQ regularizes the assignment distributions using the InfoNCE loss between node pairs. For each node, this involves computing similarities with $k$ positive and $k$ negative samples, each over $K$-dimensional distributions. The total cost is $O(|\mathcal{V}| \cdot (2k) \cdot K)$.

\myparagraph{Scalability on Large Graphs}
The model is easy to scale to large graphs, as the same strategy can be adopted for each minibatch. For example, we implement the same sampling strategy and construct the contrastive set in each input mini-batch. We additionally include the performance of RGVQ against vanilla VQ on ogbn-arxiv, ogbn-product, and ogbn-proteins in Table~\ref{tab:ogb_results}.

\end{document}